\documentclass[acmsmall,screen,nonacm]{acmart}

\setcopyright{none}
\acmDOI{} \acmISBN{} \acmPrice{}

\title{Governance by Evidence: Regulated Predictors in Decision-Tree Models}

\author{ALEXIOS VESKOUKIS}
\orcid{0009-0001-0791-4783}
\affiliation{%
  \institution{School of Science and Technology, Hellenic Open University}
  \city{Patras}
  \country{Greece}}
\email{alexios.veskoukis@ac.eap.gr}

\author{DIMITRIS KALLES}
\orcid{0000-0003-0364-5966} 
\affiliation{%
  \institution{School of Science and Technology, Hellenic Open University}
  \city{Patras}
  \country{Greece}}
\email{kalles@eap.gr}

\usepackage{graphicx}
\graphicspath{{../figures/}{figures/}} 
\DeclareGraphicsExtensions{.pdf,.png,.jpg} 
\usepackage{float}      
\usepackage{placeins}   

\usepackage[export]{adjustbox} 

\usepackage{cuted}      
\usepackage{caption}

\usepackage{needspace}  
\usepackage{etoolbox}   

\AtBeginEnvironment{figure}{\par\Needspace{12\baselineskip}}

\AfterEndEnvironment{figure}{\par\vspace{-3pt}\noindent\ignorespaces}

\usepackage{tcolorbox}
\tcbuselibrary{listings,listingsutf8,breakable,skins}
\tcbset{listing engine=listings, listing utf8=utf8}

\usepackage{listings}
\usepackage{amsfonts}

\usepackage{listingsutf8} 

\lstdefinestyle{promptstylecompact}{%
  basicstyle=\ttfamily\footnotesize,
  columns=fullflexible,
  keepspaces=true,
  showstringspaces=false,
  breaklines=true,
  inputencoding=utf8,
  extendedchars=false,
  literate=
    {≤}{{$\le$}}1
    {≥}{{$\ge$}}1
    {→}{{$\to$}}2
    {•}{{$\bullet$}}1
    {—}{{---}}3
    {–}{{-}}1
}

\newtcblisting{promptblock}[2][]{%
  listing engine=listings,
  listing utf8=utf8,
  listing only,
  listing options={style=promptstylecompact,inputencoding=utf8,columns=fullflexible,keepspaces=true},
  enhanced, breakable,
  colback=black!2,
  colframe=black!55,
  boxrule=0.3pt,
  title={#2},
  fonttitle=\bfseries\footnotesize,
  colbacktitle=black!8,
  coltitle=black,
  title filled,
  top=0.9ex,
  bottom=0.6ex,
  left=0.6mm, right=0.6mm,
  boxsep=0.6mm,
  arc=0.7mm, outer arc=0.7mm,
  before skip=4pt, after skip=4pt,
  #1
}

\usepackage{tikz}
\usetikzlibrary{positioning,arrows.meta,calc}

\begin{document}
\begin{abstract}
Decision-tree methods are widely used on structured tabular data and are valued for interpretability across many industry sectors.
However as published studies often list the predictors they used (for example age, diagnosis codes, location),
current privacy laws increasingly lead us to raise the question of how often these reported predictors fall into legally
regulated data types, and in which industries is this exposure maximum and, maybe, leading to legal risks.
In this work, we use published decision-tree papers as a practical proxy for real-world use of legally governed data types.
We compile a corpus of decision-tree studies and assign each explicitly reported predictor to a regulated data category
(for example health data, biometric identifiers, children's data, financial attributes, location traces, and government IDs),
which we then link to the specific text excerpts in European Union/United States privacy law which covers the category or the predictor itself.
Across the corpus, we observe that many reported predictors fall in regulated data categories,
with the largest shares in healthcare and clear differences across industries.
We analyze corpus-wide prevalence, industry composition, and temporal patterns of regulated predictors reported in decision-tree
studies and we assess regulation-aligned timing by marking each framework's reference year and summarizing cross-framework differences.
This work aims to map the current use of regulated data in decision-tree models so as to motivate and orient further work on
privacy-preserving methods and governance checks.
Although we focus on decision-tree studies, the evidence and governance checks that we present can inform privacy-preserving machine-learning practice more broadly.

\end{abstract}

\begin{center}
\small
This is an author version (preprint).
\end{center}

\maketitle


\section{Introduction}
Decision-tree models are widely used in applied machine learning on tabular data across domains such as healthcare,
finance, marketing, and public services. They are often chosen because their structure appears easy to inspect,
yet the predictors they use describe people, organizations, and behavior in ways that privacy and sectoral laws regulate.
Legal frameworks in the European Union and the United States define categories of regulated data and attach duties to their use,
disclosure, and retention. These legal duties motivate a closer look at how decision-tree predictors relate to law-defined
data categories across sectors and years.

Our study uses the published decision-tree literature as a proxy for model use in practice.
We assemble a corpus of tree-based papers from multiple sectors, identify the predictors that authors explicitly report,
and align these predictors with an ontology of regulated data categories anchored in specific clauses of European Union
and United States regulation. AI-assisted data extraction and human validation link predictors to paragraph-cited legal fragments.
On this basis we quantify how reported predictors are distributed across sectors and regulated categories, estimate what share
of them falls within law-defined categories, rank regulations by how often they govern the predictors and categories we observe,
and trace how the use of regulated predictors changes over time.

In this study, we first situate our work in conceptual and empirical background and related research.
We then describe how we construct the dataset of decision-tree papers, extract predictors, and map them to regulated data
categories and legal clauses. Next, we present our audit-based error assessment and population correction and report empirical
findings on regulated predictors across sectors, regulations, and time. We conclude by summarizing our data, code,
and replication resources, discussing limitations, and examining societal and policy implications.

\section{Background and Related Work}
This section provides conceptual and empirical background for our study.
We first review decision-tree models and their role in interpretable machine learning.
We then define regulated data categories and the multi-class schema we use to normalize predictors, and describe how we map them to regulatory clauses.
Next, we identify what existing surveys cover and what corpus-scale measurement gaps remain.
Finally, we summarize the resources we provide for responsible computing.

\subsection{Decision trees and interpretability}
Decision trees make their reasoning explicit: each split names a condition and each root–to–leaf path is a human-readable rule.
This transparency supports audit and documentation.
Canonical tree induction methods and widely used toolchains illustrate this transparency in practice \cite{quinlan1986induction,quinlan1993c45,kalles1996efficient,kalles2000stable,kalles2009lossless,hall2009weka,shapiro1987structured}.
Benchmark repositories and applied evaluations further motivate explicit predictor reporting in the literature \cite{uciml2017,kurgan2001knowledge}.
Because published tree studies often list the exact predictors (and sometimes rules) they used, we extract the reported predictors,
normalize them to a fixed attribute ontology, and anchor them to paragraph-cited legal categories.
This provides a precise basis to quantify when published practice touches regulated data categories and to assess implications for disclosure,
auditing, and risk across domains and time.

\subsection{Privacy regulations and regulated data categories}
Our analysis links paper-extracted predictors to paragraph-anchored legal evidence.
This design is motivated by privacy-preserving data mining work that operationalizes privacy risk and protection in terms of sensitive attributes and attribute classes \cite{agrawal2000privacy,verykios2004state,kreso2020datamining,krasadakis2020advanced}.
The pipeline maps each extracted decision-tree predictor to an attribute ontology and then to regulatory clauses using
quoted fragments keyed to article or sections of major regulation applicable in the EU (European Union) \cite{gdpr2016,eprivacy2002,nis2_2022,psd2_2015,crossborder_healthcare_2011} and the US (United States) \cite{hipaa_pl104_191,hitech_pl111_5,glba_pl106_102,coppa_uscode_ch91,ccpa_civ_code_title1815,cpra_prop24_2020};
the final validator cites the exact paragraphs, enabling audit and re-checks without external knowledge.
Regulated data categories are the bridge from raw mentioned predictors in scientific literature to law-defined categories.
By \emph{regulated data category} we mean the ontology \emph{ to which category that a predictor belongs}.
We use a fixed, closed multi-class schema (e.g., \textit{Health\_Clinical}, \textit{Demographic}, \textit{Behavioral}, \textit{Financial}, \textit{Device\_OnlineID}, \textit{Child\_Data}) to normalize heterogeneous predictors and to aggregate legal coverage at the class level.
This evidence-first linkage supports robust counts and comparisons later in the paper without inferring regulation beyond what the quoted text supports.

\subsection{Prior surveys and gaps}

Surveys in privacy and machine learning summarize protection mechanisms for data mining and access control \cite{agrawal2000privacy,verykios2004state,kreso2020datamining,krasadakis2020advanced}.
In this work, we focus on corpus-scale measurement for decision-tree studies where predictors are explicitly reported and can be linked to paragraph-cited legal text.
We treat these counts as an evidence-first proxy for reported practice across sectors and time.

To our knowledge, corpus-scale measurement remains limited.
We lack counts of how often published decision-tree papers report predictors in legally governed data types and how this varies by industry and year.
Because many deployments in practice adopt methods first introduced in the scholarly literature, and applied studies analyze
sector datasets and standard benchmarks, these counts constitute an informative proxy for uptake in industry.
Closing this gap helps quantify how much reported material is legally governed, which data types appear most often, and which industries face the greatest exposure.

Surveys in privacy-preserving data mining summarize foundations and modern research agendas \cite{agrawal2000privacy,verykios2004state,kreso2020datamining,krasadakis2020advanced}.
Existing tree-level privacy-preserving approaches include parsimonious downgrading, hiding classification rules, and decision-tree rule-hiding via data set operations and local distortion, including prototypes and case studies \cite{chang1998parsimonious,natwichai2005hiding,kalles2015hiding,kalles2016ecai,kalles2016praise,feretzakis2018ifip,feretzakis2019entropy334,feretzakis2019entropy66,feretzakis2019medical,feretzakis2019fintech,feretzakis2020setn,feretzakis2020csedu,feretzakis2022diabetes}.

\subsection{Responsible computing context and contributions}

We provide resources for responsible computing.
First, measurement: corpus-scale counts showing where reported predictors fall under law-defined data types.
Second, evidence: paragraph-cited legal fragments and item-level provenance that support each count and label.
Third, guidance: clear checks for disclosure and audit based on these measurements.
We release code, data, and prompts to enable independent review and reproduction.

\section{Dataset Construction}\label{sec:dataset-construction}

Our analysis targets a \emph{precision-focused} subset: papers whose abstracts contain at least one decision-tree predictor that our pipeline mapped to a regulated data category and that passed human validation (\texttt{Regulated+High}).
All estimates and figures are therefore conditional on inclusion in this validated subset.
Any corpus-level prevalence numbers are conservative lower bounds rather than population totals.
Each gate in the pipeline is accompanied by a stratified human \emph{audit check} used to estimate precision and adjust counts; Section~\ref{sec:audit-error-correction} specifies this audit design.

We now fix terminology used throughout the pipeline:
\begin{itemize}
\item \emph{Predictor}: an input variable reported in a study (for example age, diagnosis code, device identifier).
\item \emph{Regulated data category (RDC)}: a law-defined data type family used for mapping and reporting (for example health-clinical data, biometric data, child data, location data).
\item \emph{Regulated predictor}: a reported predictor that maps to an RDC covered by the cited legal text.
\item \emph{Industry sector}: the application area of the study (for example Banking/Finance, Healthcare, Education).
\item \emph{Named regulation}: the specific law referenced (for example GDPR, HIPAA, CCPA/CPRA).
\item \emph{Paragraph-cited legal fragment}: the exact paragraph of a law cited for coverage.
\item \emph{Audit confirmation rate}: the share of automated labels confirmed by human review.
\item \emph{Post-reference period}: the time after the regulation's reference year $E_r$.
\item \emph{Corpus}: the set of papers we analyze.
\item \emph{Gate.} In this pipeline, a \emph{gate} is a deterministic pass/fail acceptance check applied to the outputs of step $k$; only items that satisfy the step’s validation rule advance to step $k{+}1$, while others are blocked.
\end{itemize}

We fix one canonical reference year $E_r$ per regulation for all vertical markers, labels, and time--series models; $E_r$ equals the year listed in Appendix~\ref{sec:regulation-enactment-timeline}.
By default this is the statute's enactment/adoption (or entry-into-force for EU instruments); where a different milestone is used, the appendix notes it.
All captions that refer to \("\)enactment\("\) use $E_r$ as shorthand.

\newcommand{\rdc}{RDC}
\newcommand{\RDC}{Regulated Data Category}
\newcommand{\sector}{industry sector}
\newcommand{\sectors}{industry sectors}

\subsection{AI assistance in corpus construction}
We employed two widely used LLM-based capabilities: \emph{information extraction} (prompted LLM) to pull explicitly reported decision-tree predictors
and paragraph-cited statute excerpts, and \emph{text classification} (supervised/prompted classifier) to screen decision-tree relevance and assign each predictor to a regulated data category (RDC).
Outputs followed fixed schemas and versioned prompts; model metadata (name, version, parameters, timestamps) was logged; stratified human audits spot-checked accuracy.
LLM (large language model) extraction and classification can introduce systematic bias. The model may prefer certain phrasings, acronyms,
or English-only terms, which can under- or over-extract specific predictor types and shift regulated data category (RDC) counts and between-group comparisons.
We mitigate with fixed output schemas, versioned prompts, logged model metadata, and stratified human audits, but residual bias may remain and should be considered when interpreting aggregate rates.

Extraction and validation employed an OpenAI model (App.~\ref{app:ai-model-gpt41}). To enable reproduction without additional API calls,
we provide cached AI outputs (predictors, statute fragments, labels, rationales) together with prompts, schemas, and model metadata.

\subsection{Corpus retrieval and filtering}
We queried Crossref and OpenAlex for scholarly articles in a predefined set of target industry sectors \cite{crossref2024,openalex2024},
then merged and deduplicated records by DOI (Crossref 10{,}023; OpenAlex 9{,}382; total 19{,}405).
Each record retained title and abstract. An automated abstract screen removed non-relevant search hits by detecting explicit
references to decision-tree methods or clear synonyms. Records passing this screen constitute the \emph{decision-tree–relevant set (DT-relevant)}.
To validate this screening procedure, we drew a stratified sample of 1{,}000 records
across the two sources after merge and deduplication.
A human reviewer judged decision-tree relevance on this sample to estimate the precision
of the automated relevance screen.

\subsection{Relevance screening, industry assignment, and inclusion}
From the decision-tree-relevant set (8{,}386), an automated classifier assigned each article to exactly one query industry using a 12-sector taxonomy (shown below).
For downstream analyses, we retain only articles that
(i) are decision-tree relevant and
(ii) whose automated industry sector assignment matches the industry used at search time, yielding 4{,}686 records.
This agreement filter is a precision-oriented step that intersects two independent signals (search keyword retrieval and classifier assignment),
reducing false positives from either source at the cost of some recall; we quantify precision on the 1{,}000-record audit sample.

\begin{quote}\footnotesize
\begin{minipage}[t]{0.48\linewidth}
\begin{itemize}\itemsep2pt
  \item Banking/Finance
  \item Healthcare/Pharma
  \item Insurance
  \item E-commerce/Retail
  \item Telecom/Network Security
  \item Social Media
\end{itemize}
\end{minipage}\hfill
\begin{minipage}[t]{0.48\linewidth}
\begin{itemize}\itemsep2pt
  \item Education/Learning Analytics
  \item IoT/Smart Systems
  \item Government/Public Administration
  \item Cybersecurity/Intrusion Detection
  \item HR/Recruitment
  \item Transportation/Logistics
\end{itemize}
\end{minipage}
\end{quote}

Figure~\ref{fig:prisma} presents the full screening pipeline and resulting corpus composition.
The initial identification stage yielded 19{,}405 records from Crossref (10{,}023) and OpenAlex (9{,}382).
After screening for decision-tree relevance, 8{,}386 articles were retained.
Among these decision-tree relevant articles (Table~(a) in Figure~\ref{fig:prisma}), Healthcare/Pharma comprises the largest sector with 1{,}714 articles, followed by Education/Learning Analytics (992), Cybersecurity/Intrusion Detection (863), and IoT/Smart Systems (804).
Applying the industry match filter reduces the corpus to 4{,}686 articles (Table~(b) in Figure~\ref{fig:prisma}), with Healthcare/Pharma (747) and Education/Learning Analytics (631) remaining the dominant sectors.
To validate this classification step, we used the same 1{,}000-record audit sample where a human reviewer independently assigned industry sectors.
Agreement between human and automated assignments quantifies the reliability of this step.

\begin{figure}[H]
  \centering
  \scriptsize
  \setlength{\tabcolsep}{6pt}
  \renewcommand{\arraystretch}{1.12}

  \begin{tabular}{l r l}
    \hline
    \textbf{Stage} & \textbf{Count} & \textbf{Notes}\\
    \hline
    Identification & 19{,}405 & Crossref 10{,}023; OpenAlex 9{,}382\\
    Screening (AI relevance) & 8{,}386 & Excluded (Not relevant: 11{,}016; Error: 3)\\
    Eligibility (industry assignment) & 8{,}386 & Domain distribution below (a)\\
    Inclusion (industry match with query) & 4{,}686 & Per-industry inclusion below (b)\\
    \hline
  \end{tabular}
\label{fig:table1}
  \vspace{8pt}

  \begin{tabular}{l r}
    \hline
    \textbf{(a) Domain over 8{,}386 AI-relevant} & \textbf{n}\\
    \hline
    banking\_finance & 689\\
    cybersecurity\_intrusion\_detection & 863\\
    ecommerce\_retail & 431\\
    education\_learning\_analytics & 992\\
    government\_public\_admin & 399\\
    healthcare\_pharma & 1{,}714\\
    hr\_recruitment & 157\\
    insurance & 184\\
    iot\_smart\_systems & 804\\
    manufacturing\_industry & 1\\
    social\_media & 635\\
    telecom\_network\_security & 233\\
    transportation\_logistics & 625\\
    none\_of\_the\_above & 659\\
    \hline
  \end{tabular}
\label{fig:table2}
  \vspace{8pt}

  \begin{tabular}{l r}
    \hline
    \textbf{(b) Final inclusion (industry match), $n{=}4{,}686$} & \textbf{n}\\
    \hline
    banking\_finance & 405\\
    cybersecurity\_intrusion\_detection & 364\\
    ecommerce\_retail & 301\\
    education\_learning\_analytics & 631\\
    government\_public\_admin & 237\\
    healthcare\_pharma & 747\\
    hr\_recruitment & 107\\
    insurance & 163\\
    iot\_smart\_systems & 540\\
    social\_media & 542\\
    telecom\_network\_security & 159\\
    transportation\_logistics & 490\\
    \hline
  \end{tabular}
\label{fig:table3}

  \caption{Identification and screening totals; (a) industry distribution over AI-relevant records ($n{=}8{,}386$); (b) per-industry counts in the final inclusion set after requiring the AI-assigned industry to match the query industry ($n{=}4{,}686$).}
  \Description{Table (Stage counts): Identification 19{,}405 (Crossref 10{,}023; OpenAlex 9{,}382); Screening retained 8{,}386 and excluded 11{,}019; Eligibility 8{,}386; Inclusion 4{,}686. Table (a): per-industry counts over the 8{,}386 AI-relevant records. Table (b): per-industry counts for the 4{,}686 records retained after industry match with the query industry.}
  \label{fig:prisma}
\end{figure}

\subsection{Predictor extraction}
For each included abstract, we harvested candidate predictors explicitly named as inputs to a decision-tree model (e.g., “predictors,” “attributes,” “features”).
Each candidate is linked to its evidence sentence and bound to article, year, and industry.
Normalization collapses obvious synonyms while preserving the original mention for auditability.

\subsection{Predictor validation}
To ensure that extracted items are genuine decision-tree predictors rather than outcomes, labels, or other mentions, we apply a stringent validation gate that inspects each harvested item for clear abstract-level evidence of predictor status.
A separate validation pass confirmed that every harvested item is a predictor in a decision-tree context and not an outcome,
label, a generic group descriptor or a mentioned cluster. Ambiguous cases were rejected to avoid inflating downstream counts; only items with
clear abstract-level evidence were retained. This gate is conditioned on the prior inclusion set ($4{,}686$ decision tree and industry-validated articles).

Figure~\ref{fig:predictor_validation_flow} presents the validation results.
Of the 4{,}686 articles entering this stage, 596 were retained as containing valid predictors while 4{,}090 were excluded.
Among the 596 validated articles, Healthcare/Pharma accounts for the largest share with 127 articles, followed by Education/Learning Analytics (118) and Transportation/Logistics (86).
To validate this extraction step, a human verified on the audited sample that extracted items are truly decision-tree predictors named
in the abstract and not outcomes, generic group labels or any other irrelevant mention.

\begin{figure}[H]
  \centering
  \scriptsize
  \setlength{\tabcolsep}{6pt}
  \renewcommand{\arraystretch}{1.12}

  \begin{tabular}{l r l}
    \hline
    \textbf{Stage} & \textbf{Count} & \textbf{Notes}\\
    \hline
    Incoming from prior gate & 4{,}686 & Domain-validated inclusion set\\
    predictor validation: Valid & 596 & Retained for downstream steps\\
    predictor validation: Not valid & 4{,}090 & Excluded\\
    \hline
  \end{tabular}

  \vspace{8pt}

  \begin{tabular}{l r}
    \hline
    \textbf{Domain distribution within Valid ($n{=}596$)} & \textbf{n}\\
    \hline
    banking\_finance & 30\\
    cybersecurity\_intrusion\_detection & 6\\
    ecommerce\_retail & 52\\
    education\_learning\_analytics & 118\\
    government\_public\_admin & 42\\
    healthcare\_pharma & 127\\
    hr\_recruitment & 15\\
    insurance & 19\\
    iot\_smart\_systems & 56\\
    social\_media & 38\\
    telecom\_network\_security & 7\\
    transportation\_logistics & 86\\
    \hline
  \end{tabular}

  \caption{Predictor-validation counts conditioned on the industry-validated inclusion set. Top: Stage counts. Bottom: per-industry distribution among articles passing predictor validation.}
  \Description{Counts: prior gate 4{,}686; predictor validation retained 596 and excluded 4{,}090. Valid-industry distribution sums to 596 across the listed industry sectors.}
  \label{fig:predictor_validation_flow}
\end{figure}

\subsection{Regulated data category assignment}
To enable systematic analysis of which types of regulated data appear in decision tree models, we classify each validated predictor into a privacy-aware ontology of regulated data categories.
Validated predictors were mapped to a fixed privacy-aware ontology (RDC) of thirteen classes to avoid inflating regulated counts:
\begin{quote}\footnotesize
\begin{minipage}[t]{0.48\linewidth}
\begin{itemize}\itemsep2pt
  \item Identifier/PII
  \item Contact Info
  \item Device/Online ID
  \item Biometric
  \item Location/IoT
  \item Health/Clinical
  \item Financial
\end{itemize}
\end{minipage}\hfill
\begin{minipage}[t]{0.48\linewidth}
\begin{itemize}\itemsep2pt
  \item Child Data
  \item Demographic
  \item Behavioural
  \item Environmental
  \item Operational/Business
  \item Other
\end{itemize}
\end{minipage}
\end{quote}

In this step, the unit of analysis is the \emph{predictor} and a unique article (DOI); each row represents a unique predictor extracted in the prior gate.
Figure~\ref{fig:attribute_class_assignment} presents the assignment results for 1{,}749 unique predictors entering this step.
The distribution reveals that Health\_Clinical is the most frequent category with 278 predictors, followed by Behavioural (210), Environmental (184), and Child\_Data (179).
The Other category, which captures predictors not fitting the primary taxonomy, accounts for 598 predictors.
Financial (94), Demographic (86), and Operational\_Business (81) represent moderate-frequency categories, while Location\_IoT (32), Device\_OnlineID (5), and Identifier\_PII (2) are less common.
To validate this mapping step, a human assigned RDCs for the audited predictor subset to estimate agreement with the automated mapper.

\begin{figure}[H]
  \centering
  \scriptsize
  \setlength{\tabcolsep}{6pt}
  \renewcommand{\arraystretch}{1.12}

  \begin{tabular}{l r}
    \hline
    \textbf{Stage} & \textbf{Count}\\
    \hline
    Unique predictors entering this step & 1{,}749\\
    \hline
  \end{tabular}

  \vspace{8pt}

  \begin{tabular}{l r}
    \hline
    \textbf{RDCs (over $n{=}1{,}749$ predictors)} & \textbf{n}\\
    \hline
    Behavioural & 210\\
    Child\_Data & 179\\
    Demographic & 86\\
    Device\_OnlineID & 5\\
    Environmental & 184\\
    Financial & 94\\
    Health\_Clinical & 278\\
    Identifier\_PII & 2\\
    Location\_IoT & 32\\
    Operational\_Business & 81\\
    Other & 598\\
    \hline
  \end{tabular}

  \caption{Regulated data category assignment. Top: Stage count showing the number of unique predictors entering the assignment step.
  Bottom: distribution of assigned classes (classes with zero count in this step are omitted). Totals sum to $1{,}749$.}
  \Description{Unique predictors: 1{,}749.
  Class counts: Behavioural 210; Child\_Data 179; Demographic 86; Device\_OnlineID 5; Environmental 184; Financial 94; Health\_Clinical 278; Identifier\_PII 2; Location\_IoT 32; Operational\_Business 81; Other 598.}
  \label{fig:attribute_class_assignment}
\end{figure}

\subsection{Regulatory fragment catalog}
We parsed the official texts of thirteen frameworks—GDPR, ePrivacy Directive, NIS2, PSD2, EU eHealth Network, CCPA, CPRA, HIPAA, HITECH, GLBA, COPPA, FERPA, and ECPA—and segmented each document into passages using legal headings (for example “Article …” or “§ …”).
\footnote{We focus on EU and US regulations as they represent the most widely enforced
consumer privacy frameworks globally and cover the majority of the corpus jurisdictions.
This list includes major horizontal (GDPR, CCPA/CPRA) and sector-specific (HIPAA, COPPA,
FERPA) laws applicable as of 2025.}

For each passage we stored the regulation identifier, the article or section reference, and—when present—the clause or recital label.
We retained only excerpts that mention a regulated data category (RDC) or its data elements and tagged each retained excerpt with one or more RDCs based solely on its text.
These tagged excerpts are the only legal evidence used downstream; no summaries or secondary sources are used.

\subsection{Validation protocol}\label{sec:validation-protocol}
We determine regulation status for each predictor through a systematic pairing and validation procedure.
For each validated predictor with an assigned regulated data category (RDC), we create a candidate pair with every regulation that has at least one catalog passage tagged with the same RDC.
For each candidate pair we supply only: (i) the predictor name, (ii) its RDC, and (iii) the set of catalog passages for that regulation tagged with that RDC, plus any passage that explicitly names the predictor.
The validator labels the pair \texttt{Regulated} if any supplied passage clearly covers either the exact predictor or its entire RDC; otherwise \texttt{Not Regulated}.
It returns a confidence tier (\texttt{High}/\texttt{Medium}/\texttt{Low}) and a brief rationale ending with the paragraph reference(s). Ties and ambiguity default to \texttt{Not Regulated}.
We retain only \texttt{Regulated} pairs with \texttt{High} confidence for the main results.
To validate this automated procedure, a human labeled regulation status and confidence on an audited subset of predictor–regulation pairs to estimate model–human agreement at this stage.
Figure~\ref{fig:reg_validation_flow} presents the validation outcomes.

\begin{figure}[H]
  \centering
  \scriptsize
  \setlength{\tabcolsep}{6pt}
  \renewcommand{\arraystretch}{1.12}

  \begin{tabular}{l r l}
    \hline
    \textbf{Stage} & \textbf{Count} & \textbf{Notes}\\
    \hline
    predictor–regulation pairs formed & 9{,}256 & joining on regulation-identified RDCs versus predictors' identified RDCs\\
    Labeled \texttt{Regulated} & 2{,}713 & By validator \\
    Labeled \texttt{Not Regulated} & 6{,}543 & By validator \\
    \hline
  \end{tabular}

  \vspace{8pt}

  \begin{tabular}{l r}
    \hline
    \textbf{Confidence among \texttt{Regulated} ($n{=}2{,}713$)} & \textbf{n}\\
    \hline
    High & 2{,}329\\
    Medium & 384\\
    \hline
  \end{tabular}

  \caption{Validation outcomes.
  Top: Stage counts for the regulation-labeling gate.
  Forming pairs of possibly regulated predictors and the regulations that the predictor's RDC appears to be regulated by.
  Bottom: confidence distribution within pairs labeled \texttt{Regulated}. For downstream reporting, only \texttt{Regulated+High} ($n{=}2{,}329$) are carried forward.}
  \Description{Stage counts: 9{,}256 predictor--regulation pairs formed, 2{,}713 labeled \texttt{Regulated} and 6{,}543 labeled \texttt{Not Regulated}. Confidence among \texttt{Regulated}: 2{,}329 High and 384 Medium; only Regulated+High (2{,}329 pairs) are used in downstream analyses.}
  \label{fig:reg_validation_flow}
\end{figure}

\subsection{Ethical compliance and risk of harm}
No human subjects or personally identifiable datasets were used; inputs were public article text, metadata, and paragraph-cited statute excerpts. This study reports empirical measurements and engineering guidance and does not provide legal advice.

\paragraph{Potential harms and mitigations.}
Risks: misusing regulated-predictors tallies to weaken disclosure, over-reliance on automated validators,
and transferring results across industry sectors without re-audit.
Mitigations: paragraph-anchored legal evidence only, conservative validator defaults with ties set to \texttt{Not Regulated},
human-in-the-loop audits on stratified samples, and release of analysis artifacts for scrutiny.

\paragraph{Equity note.}
Some regulated data categories—\textit{Demographic} and \textit{Child\_Data}—may differentially affect groups.
We explicitly flag these categories wherever they occur and, when present, recommend a one-sentence,
sector-aware note stating intended use and safeguards. This paper does not evaluate fairness outcomes.

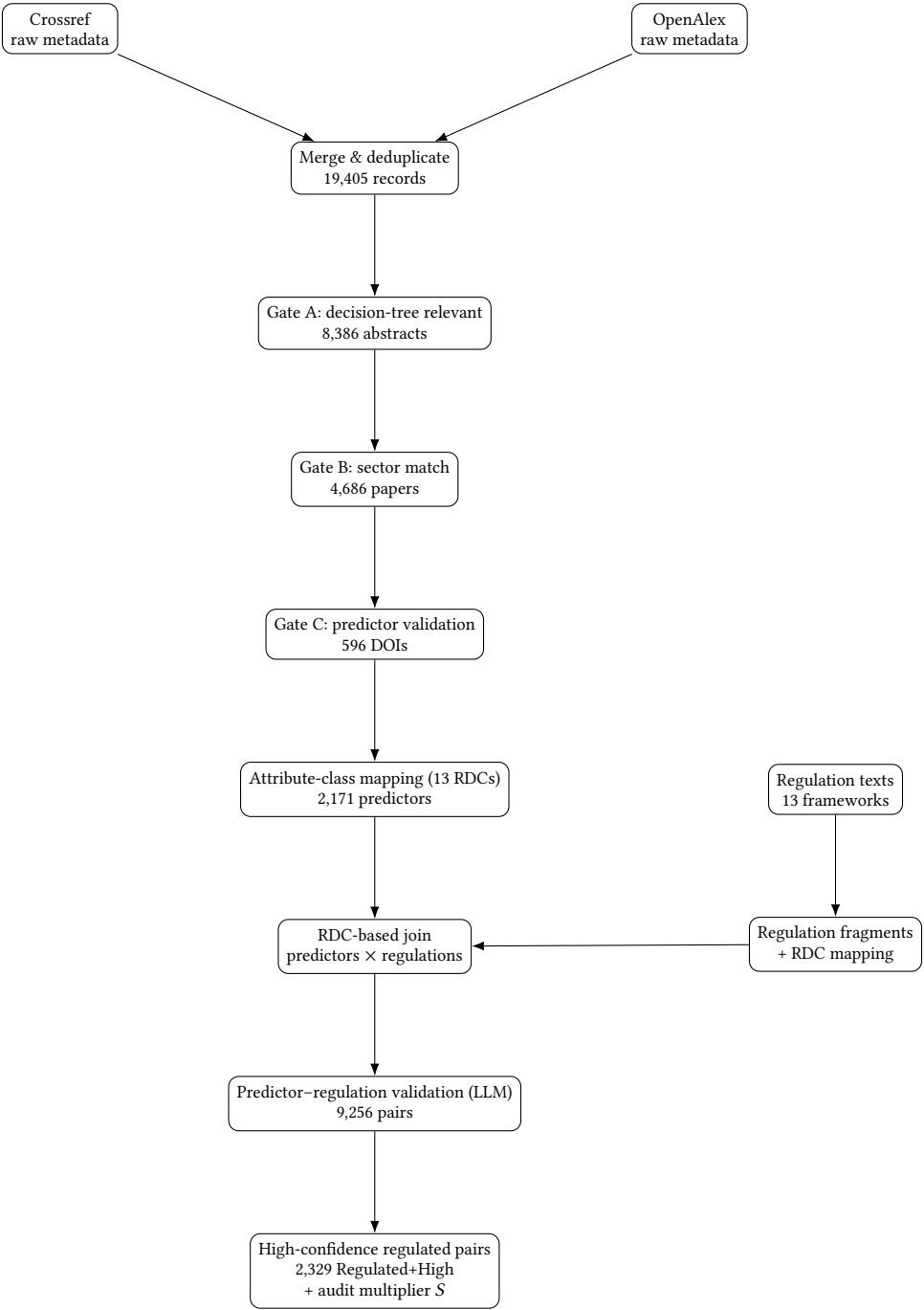
\begin{figure}[H]
  \centering
  \begin{tikzpicture}[
    node distance=1.4cm,
    every node/.style={font=\scriptsize},
    io/.style={rectangle,rounded corners,draw,align=center},
    process/.style={rectangle,rounded corners,draw,align=center},
    arrow/.style={-Latex}
  ]

    \node[process] (merge) {Merge \& deduplicate \\ 19{,}405 records};

    \node[io, above left=1.2cm and 2.4cm of merge] (crossref) {Crossref \\ raw metadata};
    \node[io, above right=1.2cm and 2.4cm of merge] (openalex) {OpenAlex \\ raw metadata};

    \node[process, below=of merge] (dt) {Gate A: decision-tree relevant \\ 8{,}386 abstracts};
    \node[process, below=of dt] (sector) {Gate B: sector match \\ 4{,}686 papers};
    \node[process, below=of sector] (predval) {Gate C: predictor validation \\ 596 DOIs};
    \node[process, below=of predval] (attr) {Attribute-class mapping (13 RDCs) \\ 2{,}171 predictors};

    \node[io, right=3.6cm of attr] (regtexts) {Regulation texts \\ 13 frameworks};
    \node[process, below=of regtexts] (regmap) {Regulation fragments \\ + RDC mapping};

    \node[process, below=of attr] (join) {RDC-based join \\ predictors $\times$ regulations};
    \node[process, below=of join] (pair) {Predictor--regulation validation (LLM) \\ 9{,}256 pairs};
    \node[process, below=of pair] (final) {High-confidence regulated pairs \\ 2{,}329 Regulated+High \\ + audit multiplier $S$};

    \draw[arrow] (crossref) -- (merge);
    \draw[arrow] (openalex) -- (merge);

    \draw[arrow] (merge) -- (dt);
    \draw[arrow] (dt) -- (sector);
    \draw[arrow] (sector) -- (predval);
    \draw[arrow] (predval) -- (attr);

    \draw[arrow] (regtexts) -- (regmap);
    \draw[arrow] (regmap) -- (join);

    \draw[arrow] (attr) -- (join);
    \draw[arrow] (join) -- (pair);
    \draw[arrow] (pair) -- (final);
  \end{tikzpicture}
  \caption{Dataset construction pipeline. The main branch builds the decision-tree corpus and predictor table; the parallel branch extracts regulation fragments and maps them to regulated data categories (RDCs). Their RDC-based join yields predictor--regulation pairs for LLM validation and auditing.}
  \label{fig:dataset_flow}
\end{figure}

\section{Audit-based Error Assessment and Population Correction}\label{sec:audit-error-correction}
This section describes how we use a stratified human audit to quantify error in the automated pipeline and adjust corpus-level counts.
We first outline the sampling design, weighting scheme, and blinding protocol used for the audit.
We then report step-specific agreement and precision metrics across relevance, industry sector, predictor validity, regulated data category, and regulation status.
Finally, we derive a conservative compound multiplier from these audits and explain how we apply it to correct population-level estimates throughout the paper.

\subsection{Sampling design and weighting}
We audited using a sample of $1{,}000$ DOIs drawn to match the source composition of the full corpus ($N{=}19{,}405$).
The corpus has two source groups: Crossref ($N_1{=}10{,}023$) and OpenAlex ($N_2{=}9{,}382$).
From each group $h\in\{1,2\}$ we drew $n_h$ records in proportion to its share $N_h/N$.
Each audited record from group $h$ was assigned the weight $w_h = N_h / n_h$.
All audit estimates use these weights to represent the full corpus. Human labels exist only for the audited records.
Automated labels are used corpus-wide; audit results supply a single compound multiplier
$S=0.415065$ applied to AI \emph{Regulated+High} predictor–regulation pair counts at reporting time.
\noindent\textit{Sample size justification.}
For a proportion $p$, the 95\% margin of error under stratified proportional allocation is well approximated by the simple–random–sampling (SRS) formula with finite–population correction (FPC):
\[
\mathrm{MOE}_{95}(p,n)
= z_{0.975}\sqrt{\frac{p(1-p)}{n}}\,
\sqrt{\frac{N-n}{\,N-1\,}},
\quad z_{0.975}=1.96.
\]
The worst case is $p=0.5$. With $N=19{,}405$ and $n=1{,}000$,
\[
\mathrm{MOE}_{95}(0.5,1000)
= 1.96\sqrt{\frac{0.25}{1000}}\sqrt{\frac{19{,}405-1{,}000}{19{,}405-1}}
= 0.03018 \ \text{(3.02\%)}.
\]
If one required $\mathrm{MOE}_{95}\le 3.00\%$ at $p=0.5$, the SRS target is
\[
n_0=\frac{z_{0.975}^2\,0.25}{0.03^2}=1{,}067.11,
\qquad
n=\frac{n_0N}{n_0+N-1}=1{,}011.54 \ \text{(with FPC to }N=19{,}405\text{)}.
\]
Thus $n=1{,}000$ is within $1.5\%$ of the FPC–adjusted target and yields a worst–case 95\% margin $\le 3.02\%$; for $p$ farther from $0.5$ the margin is smaller (e.g., $p=0.20 \Rightarrow 2.48\%$) \cite{agresti2013categorical}.

\textit{Blinding.} Human reviewer was blind to the AI labels and rationales for all audited items; only the source text, predictor name, and legal fragments were shown.

\subsection{Step-specific audit findings}

\paragraph{Decision tree relevance.}
Chance-corrected inter-rater reliability (Cohen's $\kappa$, AI vs.\ human) = 0.938529 \cite{cohen1960coefficient},
with observed agreement = 0.970270 and expected-by-chance = 0.516359 (weighted).
Weighted confusion (AI vs.\ human): TP = 7{,}639.846, FP = 480.093, FN = 96.158, TN = 11{,}166.809.
Precision of AI “Relevant” = 0.940875; miss rate among AI “Not relevant” = 0.008538.

\begin{table}[t]
  \centering
  \scriptsize
  \setlength{\tabcolsep}{8pt}
  \renewcommand{\arraystretch}{1.12}
  \caption{Relevance audit confusion matrix (weighted counts).}
  \begin{tabular}{lcc}
    \hline
    & \textbf{Human: Relevant} & \textbf{Human: Not relevant} \\
    \hline
    \textbf{AI: Relevant}     & $7{,}639.846$ (TP) & $480.093$ (FP) \\
    \textbf{AI: Not relevant} & $96.158$ (FN)      & $11{,}166.809$ (TN) \\
    \hline
  \end{tabular}
  \label{tab:audit_relevance_cm}
\end{table}

\paragraph{Industry sector assignment.}
Chance-corrected inter-rater reliability (Cohen's $\kappa$, AI vs.\ human) on all audited AI-relevant rows is
$\kappa=0.886858$ with observed agreement $P_o=0.898658$ and expected-by-chance $P_e=0.104295$ (weighted).
Among items that passed the AI industry gate ($\texttt{industry\_validated}=\texttt{industry\_searched}$),
$\kappa=0.914191$ with $P_o=0.923450$ and $P_e=0.107905$ (weighted).
For gate-pass purity diagnostics, the human agreement rate is $0.923450$ and the pass-wrong rate is $0.076550$ (weighted).

\paragraph{Predictor-validation audit confusion matrix (weighted counts).}
Chance-corrected inter-rater reliability (Cohen's $\kappa$, AI vs.\ human) on AI-gated audited DOIs is
$\kappa=0.713029$ with observed agreement $P_o=0.944573$ and expected-by-chance $P_e=0.806854$ (weighted).
Weighted confusion (AI vs.\ human): $\mathrm{TP}=375.675$, $\mathrm{FP}=118.631$, $\mathrm{FN}=139.771$, $\mathrm{TN}=4{,}027.922$.
Precision of AI ``Valid'': $0.760005$; recall vs.\ human: $0.728834$.

\begin{table}[t]
  \centering
  \scriptsize
  \setlength{\tabcolsep}{8pt}
  \renewcommand{\arraystretch}{1.12}
  \caption{predictor-validation audit confusion matrix (weighted counts).}
  \begin{tabular}{lcc}
    \hline
    & \textbf{Human: Valid} & \textbf{Human: Not valid} \\
    \hline
    \textbf{AI: Valid}     & $375.675$ (TP) & $118.631$ (FP) \\
    \textbf{AI: Not valid} & $139.771$ (FN) & $4{,}027.922$ (TN) \\
    \hline
  \end{tabular}
  \label{tab:audit_predictor_cm}
\end{table}

\paragraph{Regulated data category labeling.}
Chance-corrected inter-rater reliability (Cohen's $\kappa$, AI vs.\ human) on audited, AI-gated features is
$\kappa=0.738245$ with observed agreement $P_o=0.800000$ and expected-by-chance $P_e=0.235926$ (weighted).
For reference, the weighted class match rate is $0.800000$ over $2{,}171.0$ audited feature-weights.

\paragraph{Regulation status.}
Chance-corrected inter-rater reliability (Cohen's $\kappa$, AI vs.\ human) for status-only
(Regulated vs.\ Not Regulated) on audited, AI-gated pairs is
$\kappa=0.709241$ with observed agreement $P_o=0.895131$ and expected-by-chance $P_e=0.639327$ (weighted).
For the binary ``Regulated+High'' indicator, $\kappa=0.673886$ with $P_o=0.887640$ and $P_e=0.655459$ (weighted).
We also report the performance diagnostic for the AI Regulated+High set: $\textit{feature\_regulated\_precsion}=0.785714$
based on a weighted denominator of $1{,}941.333333$ and agreement mass $1{,}525.333333$.

\paragraph{Final propagated correction.}
We conservatively correct the AI ``Regulated+High'' total by multiplying stage-wise precision estimates.

\begin{flalign*}
& A_1:\ \text{relevance correct} &&\\
& A_2:\ \text{domain correct} &&\\
& A_3:\ \text{predictor validity correct} &&\\
& A_4:\ \text{RDC labeling correct} &&\\
& A_5:\ \text{regulation status correct} &&
\end{flalign*}

\begin{flalign*}
& P\!\left(
  A_1 \wedge A_2 \wedge A_3 \wedge A_4 \wedge A_5
  \,\middle|\,
  \mathrm{AI\text{-}RegHigh}
\right)
=
\prod_{k=1}^{5}
P\!\left(
  A_k
  \,\middle|\,
  \mathrm{AI\text{-}RegHigh},\, A_1,\ldots,A_{k-1}
\right)
= \text{overall\_multiplier}\, . &&
\end{flalign*}

Formally, let \(A_k\) denote correctness at audit stage \(k\).
For a pair with \(\mathrm{AI\text{-}RegHigh}\), the probability that all stages are correct satisfies
\[
P(A_1 \wedge \dots \wedge A_5 \mid \mathrm{AI\text{-}RegHigh})
= \prod_{k=1}^{5} P(A_k \mid \mathrm{AI\text{-}RegHigh}, A_1,\dots,A_{k-1}),
\]
by the chain rule of probability \citep[e.g.,][]{billingsley1995probability}.
Our design-weighted stage-wise audit estimates plug into this decomposition, yielding a compound multiplier \(S\), a plug-in estimate of
\(P(A_1 \wedge \dots \wedge A_5 \mid \mathrm{AI\text{-}RegHigh})\),
that we interpret as the predictive value of the AI Regulated+High label under our audit design.
We then estimate the number of truly Regulated+High pairs by multiplying automated counts by \(S\), analogous to standard misclassification and measurement-error corrections that use predictive values as multiplicative adjustments to observed totals \citep[e.g.,][]{carroll2006measurement,lash2009quantitative,pepe2003statistical}.

\begin{flalign*}
& \mathbb{E}\!\left[T_{\mathrm{corr}} \,\middle|\, \mathrm{AI\text{-}RegHigh}\right]
=
T_{\mathrm{AI\text{-}High}} \times \text{overall\_multiplier}\, . &&
\end{flalign*}

\begin{flalign*}
& T_{\mathrm{corr}}
= T_{\mathrm{AI\text{-}High}}\cdot
\underbrace{\big(\text{prec}^{\text{relevance}}\cdot \text{prec}^{\text{domain}}\cdot
\text{prec}^{\text{predictor}}\cdot \text{match}^{\text{RDC}}\big)}_{\text{upstream multiplier}}
\cdot \text{prec}^{\text{status}} \, . &&
\end{flalign*}

With \(T_{\mathrm{AI\text{-}High}}=2{,}329\), \(\text{prec}^{\text{relevance}}=0.940875\),
\(\text{prec}^{\text{domain}}=0.923450\), \(\text{prec}^{\text{predictor}}=0.760005\),
\(\text{match}^{\text{RDC}}=0.800000\), and \(\text{prec}^{\text{status}}=0.785714\),
the upstream multiplier is \(0.528265\) and the overall multiplier is \(0.415065\),
yielding \(T_{\mathrm{corr}}=966.687\) (weighted).

\subsection{Population correction used in this paper}
We report counts for the final set defined as predictor–regulation pairs that the automated validator labeled
\emph{Regulated} with \emph{High} confidence. To conservatively correct these automated totals, we multiply every
AI \emph{Regulated+High} pair count by a single compound multiplier.

\begin{flalign*}
& S=\text{prec}^{\text{relevance}}\cdot \text{prec}^{\text{domain}}\cdot
\text{prec}^{\text{predictor}}\cdot \text{match}^{\text{RDC}}\cdot \text{prec}^{\text{status}}
=0.415065\, . &&
\end{flalign*}

This scalar comes from the audit (weighted) and chance-corrected comparisons (Cohen's $\kappa$) reported above.
We apply \(S\) uniformly across breakdowns (e.g., regulation, sector, year) for pair-level totals.
Metrics defined over \emph{unique papers} are not scaled by \(S\).

\noindent\emph{Note.} We do not report $\kappa$ on conditional purity subsets (AI Regulated{+}High or AI Regulated),
because the AI label is fixed by construction, which yields \(P_o=P_e\) and thus \(\kappa=0\) identically; purity on
those subsets is a performance diagnostic, not a reliability measure.

\textbf{Estimator for confirmation.}
We use stage-wise audit estimators with design weights. Let
\(\text{prec}^{\text{relevance}}\), \(\text{prec}^{\text{domain}}\),
\(\text{prec}^{\text{predictor}}\), and \(\text{match}^{\text{RDC}}\)
be the weighted estimates from Section~3.2 (relevance, domain, predictor, RDC).
Let the final-stage confirmation for regulation \(r\) be
\(\phi_r = P(\text{Human Regulated+High}\mid \text{AI Regulated+High}, r)\).
Its audit-wide counterpart is
\[
\phi = \frac{\sum_i w_i \,\mathbb{1}[\text{Human Regulated+High}] }{\sum_i w_i}
\]
computed over audited pairs with \(\text{AI Regulated+High}\).
From the audit, \(\text{prec}^{\text{relevance}}{=}0.940875\),
\(\text{prec}^{\text{domain}}{=}0.923450\),
\(\text{prec}^{\text{predictor}}{=}0.760005\),
\(\text{match}^{\text{RDC}}{=}0.800000\), and
\(\phi{=}0.785714\).
Define the upstream multiplier
\[
m_{\text{other}}=\text{prec}^{\text{relevance}}\cdot\text{prec}^{\text{domain}}\cdot\text{prec}^{\text{predictor}}\cdot\text{match}^{\text{RDC}}=0.528265\, .
\]

\textbf{Adjustment.}
Let \(N^{\mathrm{auto,Regulated,High}}\) be the automated total of predictor--regulation pairs,
and \(N^{\mathrm{auto,Regulated,High}}[r]\) the total for regulation \(r\).
We apply a single conservative compound multiplier
\[
S \;=\; m_{\text{other}}\cdot \phi \;=\; 0.415065,
\]
and (when available) regulation-specific multipliers
\[
S_r \;=\; m_{\text{other}}\cdot \phi_r \, .
\]
Adjusted counts are
\[
N^{\mathrm{adj}} = S \, N^{\mathrm{auto,Regulated,High}},\qquad
N^{\mathrm{adj}}[r] = S_r \, N^{\mathrm{auto,Regulated,High}}[r],
\quad\text{with }S_r{=}S\text{ if }\phi_r\text{ is unavailable.}
\]
This removes the expected over-count in the automated final set; it does not add pairs missed upstream.
We do not scale metrics defined over \emph{unique papers}.

\section{Results}
Each subsection pairs a short description with an interpretation and a responsible-computing takeaway (disclosure, auditing, or policy signal).
We report each result with an explicit governance lens: what the evidence implies for disclosure practices and audit triggers for regulated attributes.
Findings characterize published practice and regulation-linked predictors; they do not provide legal advice or deployment prescriptions.
We use paragraph-anchored legal fragments, conservative labeling, and audit-calibrated estimates to limit over-claiming.

\subsection{Regulated predictor instances per regulation}
\begin{figure}[htbp]
  \centering
  \includegraphics[width=0.75\linewidth,keepaspectratio]{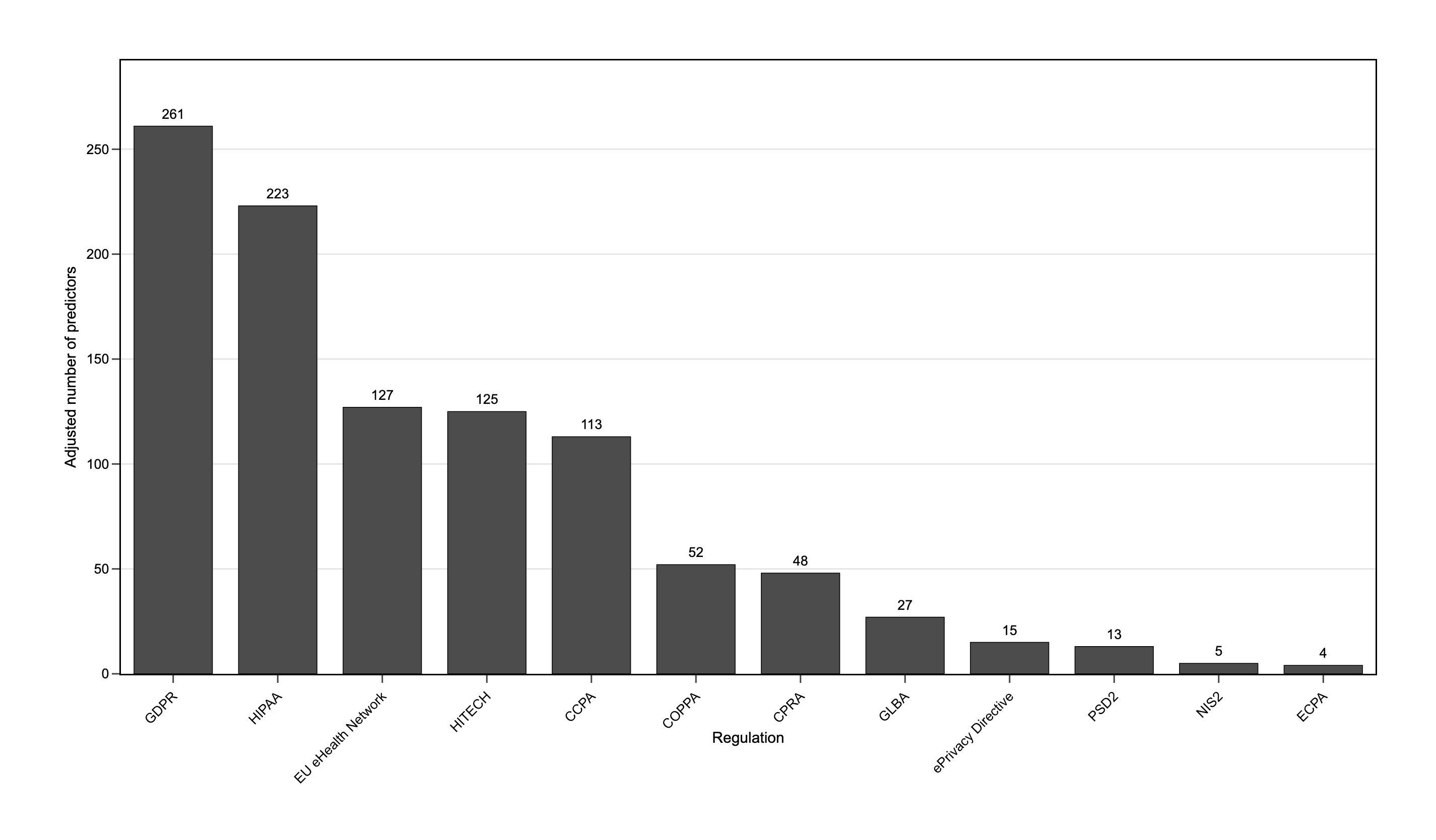}
  \caption{Adjusted number of regulated predictors per regulation.
           Counts are adjusted and bars are ordered by decreasing totals.}
  \Description{A vertical bar chart with one bar per regulation. Gray bars with
               black outlines. Numeric labels above each bar. X-axis shows
               regulation identifiers angled for readability. Y-axis shows
               adjusted counts from zero upward.}
  \label{fig:fig1}
\end{figure}

Figure~\ref{fig:fig1} presents counts of predictor–regulation pairings in the corpus
labeled with High confidence.
A higher bar indicates more uses of regulated predictors associated with that regulation
across the corpus.

The results show that General Data Protection Regulation (GDPR) and Health Insurance Portability and Accountability Act (HIPAA) dominate, indicating they govern a large share of predictors. Newer laws such as Network and Information Security 2 (NIS2) have fewer entries due to recent enactment, while sector-specific statutes such as Family Educational Rights and Privacy Act (FERPA) and Children's Online Privacy Protection Act (COPPA) cover narrower industry sectors.

\subsection{Distinct regulated predictors per regulation}

\begin{figure}[htbp]
  \centering
  \includegraphics[width=0.75\linewidth,keepaspectratio]{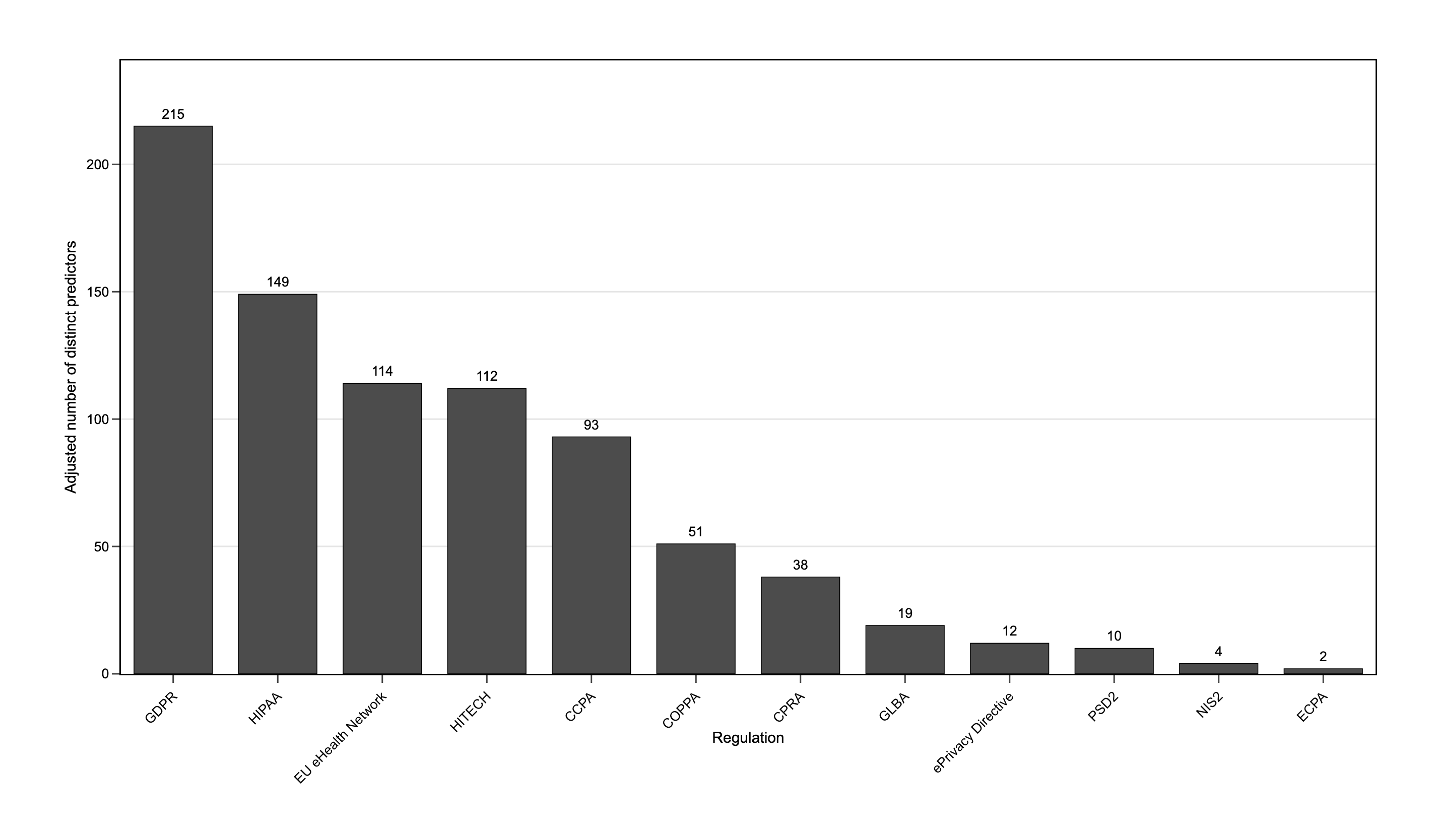}
  \caption{Adjusted number of distinct regulated predictors per regulation.
           Counts are adjusted and bars are ordered by decreasing totals.}
  \Description{A vertical bar chart with one bar per regulation. Gray bars with
               black outlines. Numeric labels above each bar. X-axis shows
               regulation identifiers angled for readability. Y-axis shows
               adjusted counts from zero upward.}
  \label{fig:fig2}
\end{figure}

While Figure~\ref{fig:fig1} captures the overall volume of regulated predictor usage, examining the diversity of unique predictors provides additional insight into the breadth of regulatory coverage.
Figure~\ref{fig:fig2} presents counts of unique predictors governed by each regulation in the high-confidence corpus.

These findings indicate that the pattern mirrors Figure~\ref{fig:fig1}: GDPR and HIPAA exhibit the broadest coverage of unique predictors, suggesting broad obligations across varied data types, whereas Electronic Communications Privacy Act (ECPA) and COPPA apply to more limited sets.

\subsection{Industry sector distribution}
\begin{figure}[htbp]
  \centering
  \includegraphics[width=0.75\linewidth,keepaspectratio]{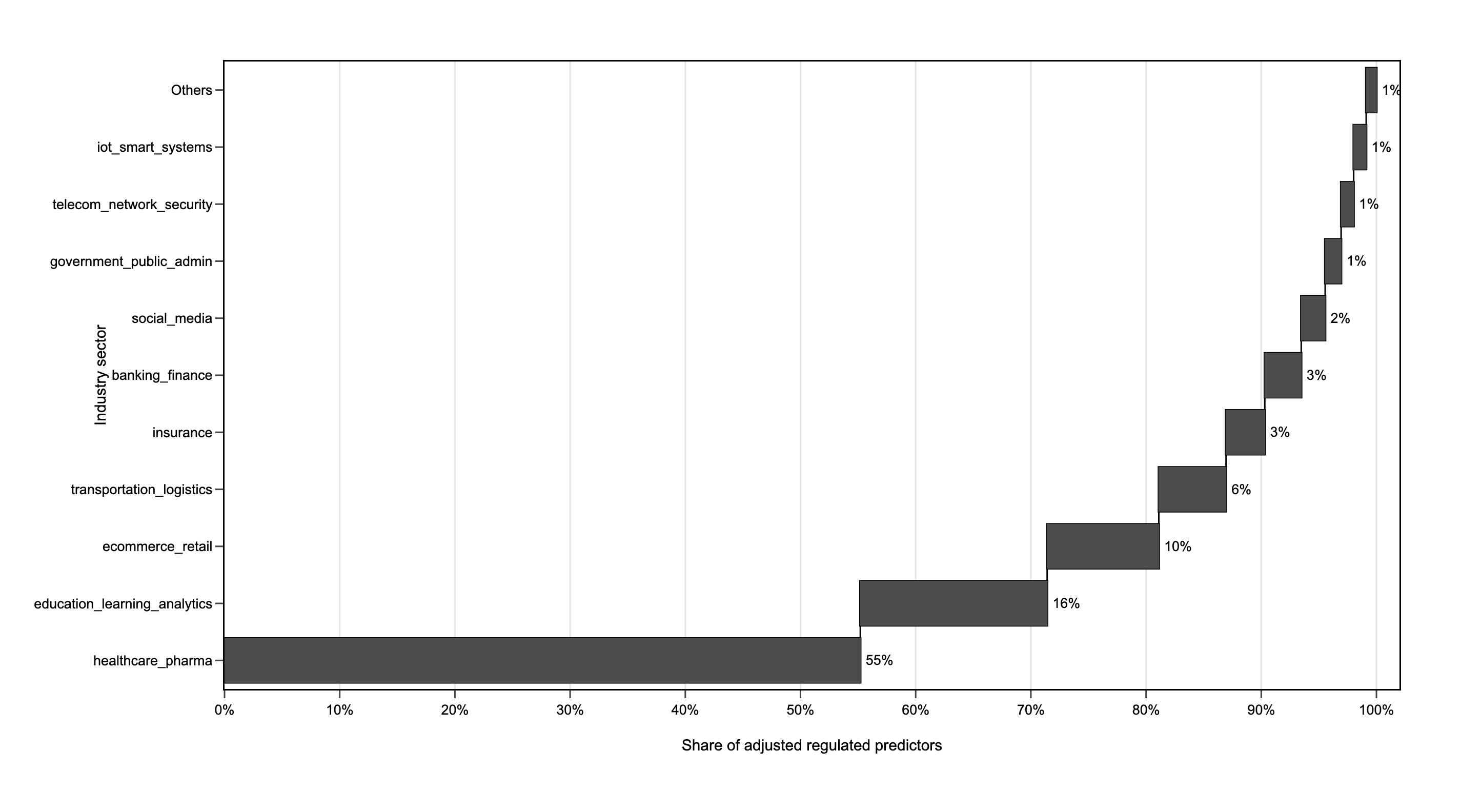}
  \caption{Industry share of adjusted regulated predictors. Top ten industry
           sectors are shown; remaining sectors are grouped as “Others.”
           Shares are computed from purity-adjusted counts.}
  \Description{A stacked horizontal bar chart with one bar segment per
               industry sector. Segments accumulate from 0 to 100\%.
               Gray bars with black outlines. Percentage labels appear at the
               end of each segment. X-axis shows 0–100\% with ticks every 10\%.}
  \label{fig:domain_distribution_records}
\end{figure}

Having examined which regulations govern the most predictors, we now examine how
regulated predictor usage is distributed across application domains.
Figure~\ref{fig:domain_distribution_records} summarizes the industry-sector share of adjusted regulated predictors (top ten sectors shown; remaining sectors grouped as “Others”), computed from purity-adjusted counts.

These findings reveal that the healthcare and pharmaceutical sector dominates, accounting for more than half of the adjusted regulated predictors.
E-commerce/retail, education, transportation, and social-media industry sectors contribute smaller but noticeable shares.
Financial services, insurance, and remaining sectors appear at only a few percent, indicating a strong skew toward health-related use cases.

\subsection{Regulated data category distribution}
\begin{figure}[htbp]
  \centering
  \includegraphics[width=0.75\linewidth,keepaspectratio]{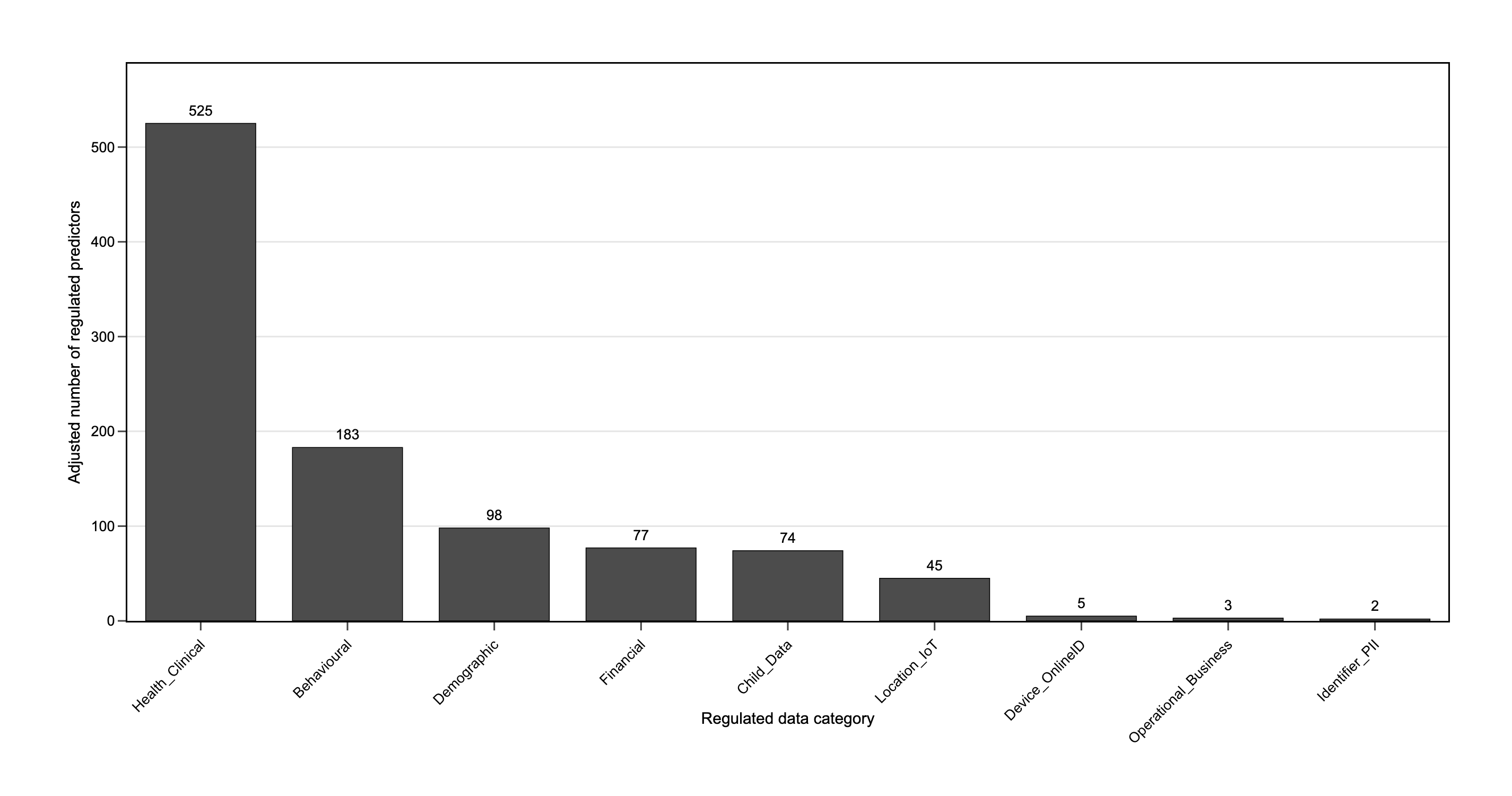}
  \caption{Regulated data category distribution showing adjusted counts of
           regulated predictors. Bars are adjusted and ordered by
           decreasing totals.}
  \Description{A vertical bar chart with one bar per regulated data category.
               Bars are gray with black outlines. Numeric labels above each bar.
               Y-axis shows adjusted counts from zero upward.}
  \label{fig:attr_class_records}
\end{figure}

Beyond industry distribution, understanding which types of data are most frequently regulated provides insight into the nature of regulatory obligations.
Figure~\ref{fig:attr_class_records} presents the number of regulated predictors falling into each RDC.

The distribution shows that Health\_Clinical attributes form the largest share, reflecting the corpus skew toward healthcare.
Behavioural and Demographic follow. Location\_IoT and Financial appear at moderate levels.
Identity-, contact-related, and operational RDCs are less frequent.

\subsection{Industry sector × regulation matrix — distinct regulated predictors}
\begin{figure}[htbp]
  \centering
  \includegraphics[width=0.75\linewidth, keepaspectratio]{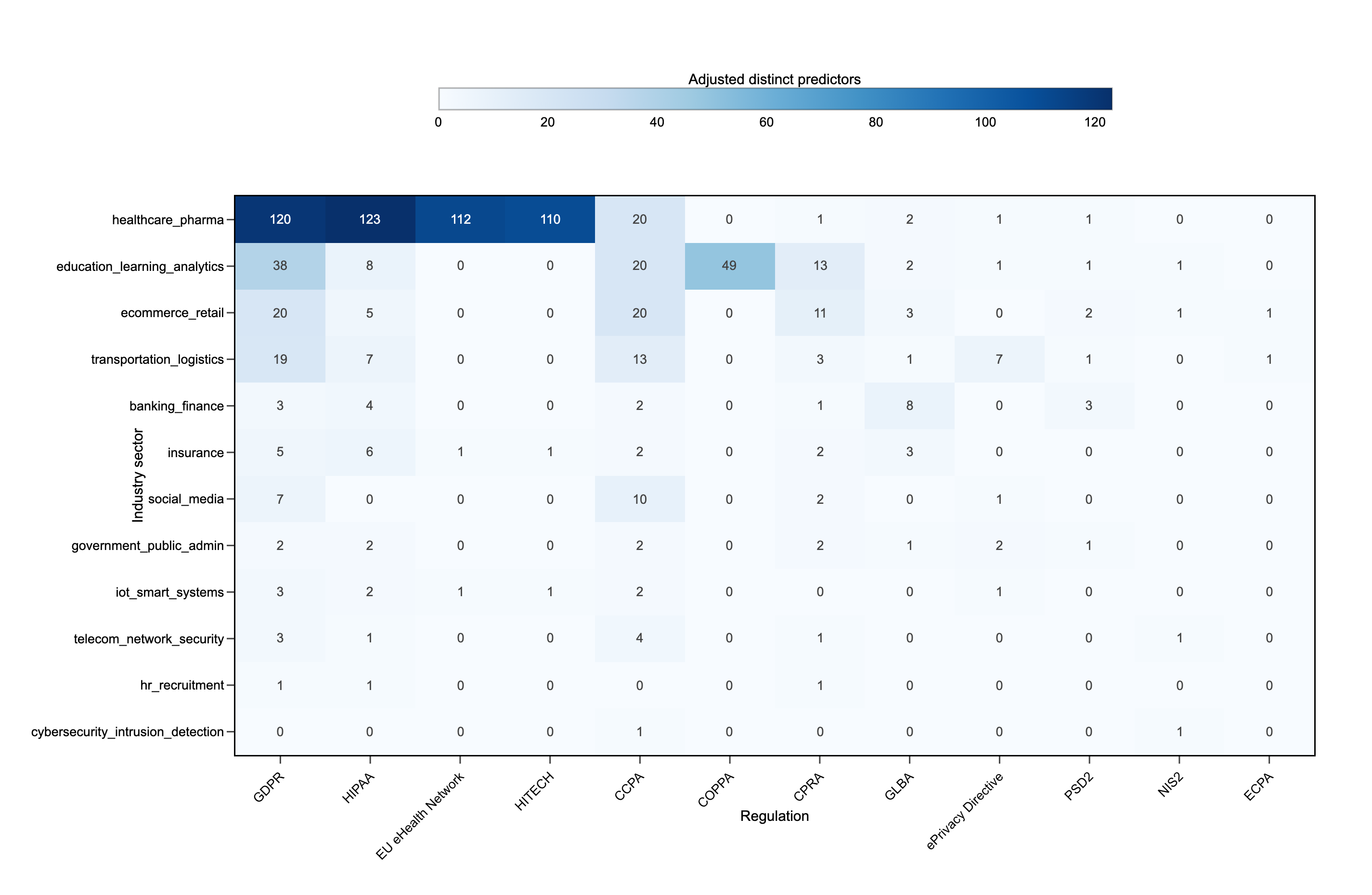}
  \caption{Industry sector by regulation heatmap showing adjusted distinct
           regulated predictors. Cell values represent the number of unique
           predictors (adjusted) discussed for each industry-regulation
           combination. Color intensity indicates higher counts. Rows and columns are ordered by decreasing
           total predictor counts.}
  \Description{A heatmap matrix with industry sectors on rows and regulations
               on columns. Cells are colored using Viridis scale from purple (low)
               to yellow (high), with white numeric values. Horizontal colorbar
               above shows color-to-value mapping.}
  \label{fig:domain_by_reg}
\end{figure}

Having examined industry sectors and regulations independently, we now analyze their intersection to identify domain-specific regulatory exposure patterns.
Figure~\ref{fig:domain_by_reg} presents each cell reporting the number of distinct predictors validated as Regulated with high confidence for each industry sector × regulation pair.
Rows and columns are ordered by their totals to reveal concentration.

The heatmap reveals that concentration is strongest in Healthcare/Pharma, led by HIPAA (297), European Union eHealth Network (269),
Health Information Technology for Economic and Clinical Health Act (HITECH) (265), and GDPR (288).
Education/Learning Analytics is next, dominated by COPPA (119), with GDPR (91), California Consumer Privacy Act (CCPA) (47),
and California Privacy Rights Act (CPRA) (31). GDPR spans many industry sectors with non-trivial counts in Transportation/Logistics (46),
E-commerce/Retail (49), and Social Media (17). CCPA/CPRA concentrate in E-commerce/Retail and Education. Gramm–Leach–Bliley Act (GLBA)
is mainly in Banking/Finance (19) with minimal Payment Services Directive 2 (PSD2). ePrivacy shows a pocket in Transportation/Logistics (18).
NIS2 and ECPA are sparse, with many zero or near-zero cells elsewhere.

%
\subsection{Yearly mean count of adjusted distinct regulated predictors per publication}
\label{sec:yr-mean-per-publication}

To examine whether regulatory enactment is associated with changes in reported predictor usage over time, we construct a time-series metric that accounts for variations in publication volume across years.
We begin by defining the metric and its components formally.
Let $r$ denote a regulation and $y$ a publication year. Let $N_y$ be the number of unique publications in the corpus in year $y$.
A \emph{regulated predictor} is a reported predictor that maps to a law-defined data category covered by $r$ under the paper's legal taxonomy.
Let $D_{y,r}$ be the set of \emph{distinct} regulated predictors (distinct by normalized name) reported across all publications in year $y$ that map to $r$.
Let $S=0.415065$ be the uniform audit-based compound multiplier applied to final AI \textit{Regulated+High} counts (Section~3.2).
Let $E_r$ be the reference year of regulation $r$ (defined in Section~2).

The yearly mean count per publication for $(y,r)$ is
\[
m_{y,r} \;=\; \frac{S\,|D_{y,r}|}{N_y},
\]
with units "adjusted distinct regulated predictors per publication."
This metric $m_{y,r}$ represents an average per publication in year $y$.
Because $N_y$ varies across years, $m_{y,r}$ is comparable over time and is not driven by publication volume.
Values above $1$ indicate that, on average, multiple such predictors are reported per publication.

\begin{figure}[htbp]
  \centering
   \includegraphics[width=0.85\linewidth, keepaspectratio]{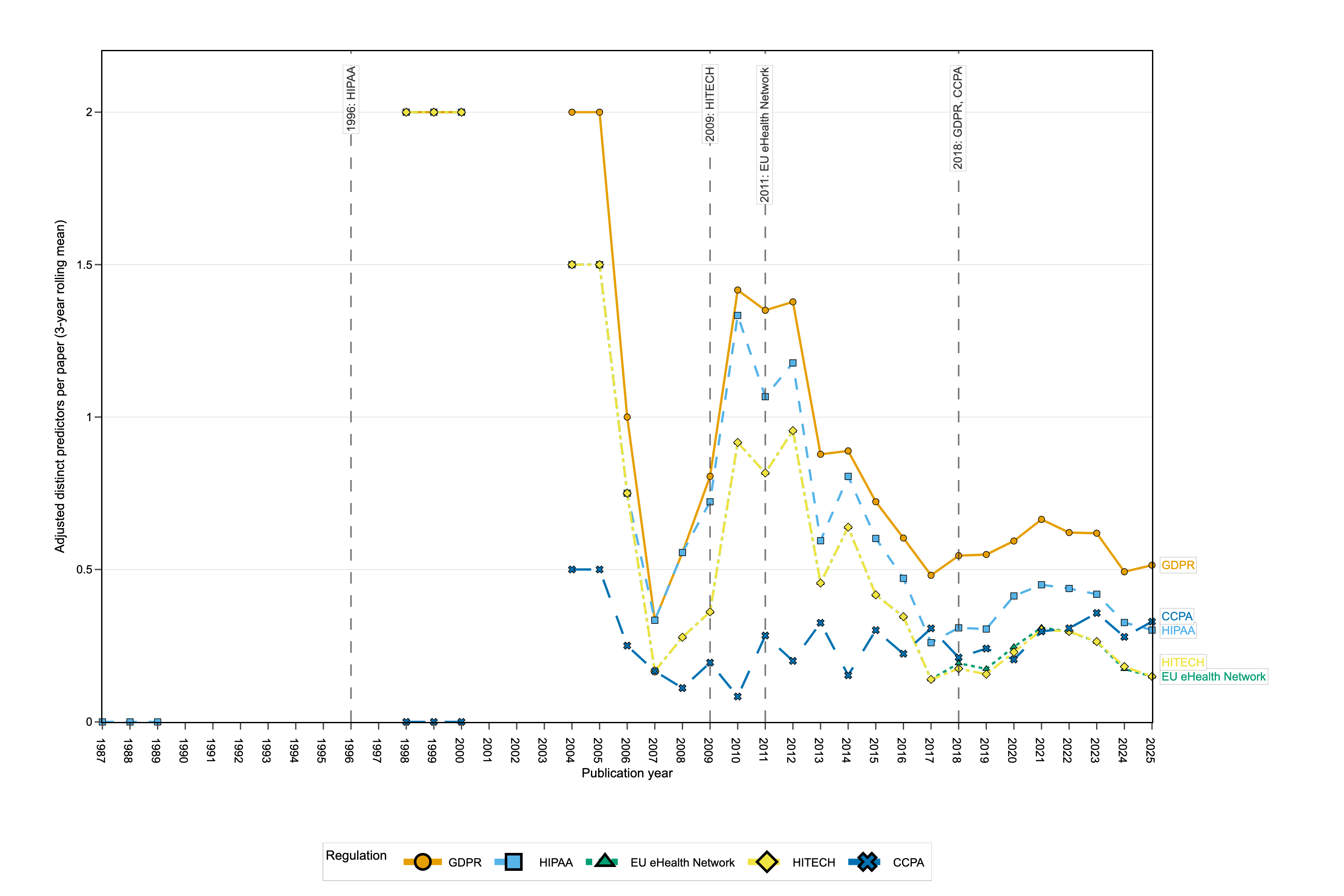}
  \caption{Per-paper rate of adjusted distinct regulated predictors by regulation,
           1987--2025 (3-year rolling mean). Five major regulations are shown using
           distinct colors, line styles, and
           marker shapes for triple redundancy. Dashed vertical lines indicate
           reference years ($E_r$) with regulation names. End labels identify each regulation
           on the right side.}
  \Description{A line chart showing trends over time from 1987 to 2025 with five
               distinct lines representing different regulations. Each line uses
               a unique combination of color, dash pattern, and marker symbol.
               The y-axis shows adjusted distinct predictors per paper, the x-axis
               shows publication year, and vertical dashed lines mark regulatory
               reference years ($E_r$) with rotated text labels. Regulation names appear
               as end labels on the right edge.}
  \label{fig:time_series_enactment}
\end{figure}

Figure~\ref{fig:time_series_enactment} presents $m_{y,r}$ plotted against year $y$ for each displayed regulation $r$.
Vertical dashed lines mark $E_r$.
Axes show integer years on the horizontal axis and $m_{y,r}$ on the vertical axis.

To assess changes at and after $E_r$, we fit an interrupted time--series (ITS) segmented regression on the \emph{counts}
with a log exposure offset for $N_y$ so that the model estimates \emph{rates per publication} within
a Poisson GLM framework \cite{camerontrivedi2013,wagner2002segreg,bernal2017its}.
Let $\text{rel}_{y,r}=y-E_r$ and $\text{post}_{y,r}=\mathbb{I}[y\ge E_r]$.
For outcome $Y_{y,r}=S\,|D_{y,r}|$ we estimate

\begin{equation}
\log E[Y_{y,r}] = \alpha_r + \beta_1\,\text{rel}_{y,r}
  + \beta_2\,\text{post}_{y,r}
  + \beta_3\,(\text{rel}\times\text{post})_{y,r}
  + \log N_y,
\label{eq:its-poisson}
\end{equation}

where $\alpha_r$ are regulation fixed effects and $\log N_y$ is the offset.
Coefficients are reported as rate ratios (RR) via $\exp(\cdot)$.
Cluster–robust standard errors are computed by regulation.
As a population–average robustness check under serial correlation, we also fit a Poisson GEE with AR(1) working correlation using the same formula and offset \cite{liangzeger1986,statsmodelsgee}.
\emph{Note:} Multiplying counts by a constant $S$ shifts the intercept but leaves slope and level-change RRs unchanged.

The statistical analysis yields the following results.
For the immediate level change at $E_r$, we obtain $\beta_2$ RR $=1.058$, 95\% CI $[0.894,\,1.252]$ (GLM); GEE $p=0.481$.
The pre--$E_r$ slope shows RR per year $=0.986$, 95\% CI $[0.964,\,1.008]$ (GLM).
The post--$E_r$ slope yields RR per year $=0.958$, 95\% CI $[0.935,\,0.981]$ (GLM), with slope--change $p=0.133$ (GLM) and $p=0.108$ (GEE).
The five--year implication from the post--$E_r$ slope is $\exp\{5(\beta_1+\beta_3)\}=0.808$ (about $19.2\%$ lower five years after $E_r$ relative to the $E_r$ year).
Model diagnostics indicate a Pearson overdispersion factor $\phi=1.068$ ($\chi^2/\text{df}$).

Interpreting these findings in the context of the figure, the plotted lines show $m_{y,r}$ by year with vertical dashed lines marking reference years $E_r$.
We read the figure using the segmented regression on counts with a $\log N_y$ exposure, which estimates rates per publication.
The results indicate three main patterns.
First, there is no immediate level change at $E_r$ (RR $=1.058$, 95\% CI $[0.894,\,1.252]$; GEE $p=0.481$).
Second, before $E_r$ the rate is flat to slightly declining (RR per year $=0.986$, 95\% CI $[0.964,\,1.008]$).
Third, after $E_r$ the decline is steeper (RR per year $=0.958$, 95\% CI $[0.935,\,0.981]$), with the slope change near significance in GEE ($p=0.108$).
Over five post--$E_r$ years this implies roughly $19.2\%$ lower values than at $E_r$ ($0.808$).
Because the model includes $\log N_y$ as an exposure, these are \emph{rate} effects and not artifacts of changes in the number of publications per year.

Several caveats apply to this analysis.
Early years have small $N_y$, increasing variance.
Right-edge years are truncated at 2025.
Concurrent domain or benchmark shifts can affect $m_{y,r}$ independently of $E_r$; the ITS design addresses timing but not mechanisms.


\subsection{Industry sector share over time}

\begin{figure}[htbp]
  \centering
  \includegraphics[width=0.8\linewidth, keepaspectratio]{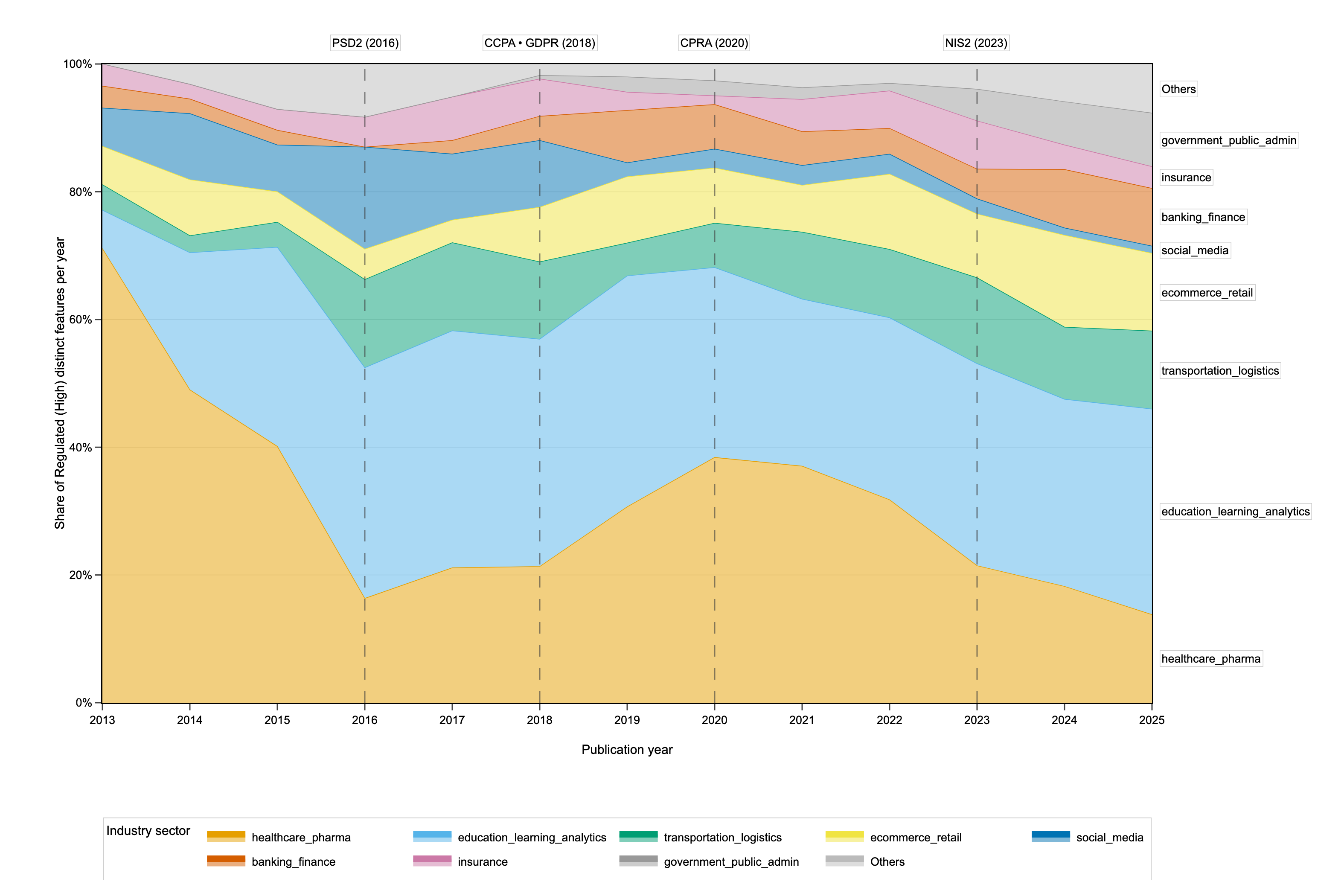}
  \caption{Annual composition of Regulated (High) distinct features by industry sector.
           Stacks sum to 100\% per year. Eight major sectors are shown; remaining sectors are grouped as \emph{Others}.
           End labels identify sectors at the right edge. Dashed lines indicate reference years ($E_r$) configured for this figure.}
  \Description{Stacked area chart from 2013 to 2025. The y-axis shows 0–100\% share per year. Colors indicate industry sectors.
               Top-8 sectors are labeled on the right edge. Years with low evidence are excluded.
               Three-year centered smoothing is applied. Vertical dashed lines show reference years ($E_r$).}
  \label{fig:domain_share_time}
\end{figure}

While the previous analyses examined aggregate patterns across the full time period, understanding how industry composition has evolved over time provides insight into shifting applications of regulated predictors.
Figure~\ref{fig:domain_share_time} presents a stacked area chart showing the per-year share of Regulated (High-confidence) distinct features by industry sector.
The denominator in year $t$ is the sum of Regulated (High) distinct features across all sectors in year $t$.
Years with fewer than 15 raw distinct items are excluded.
A 3-year centered rolling mean smooths the series.
Dashed vertical lines mark configured reference years ($E_r$).

The temporal patterns reveal several notable trends.
From 2013–2015, \emph{healthcare\_pharma} holds the largest share, then drops to a trough around 2016.
It rebounds through 2019–2020 and declines after 2021, reaching low-teens by 2025.
\emph{education\_learning\_analytics} rises from 2016, peaks around 2019–2021, and is the largest share again from 2023–2025.
\emph{transportation\_logistics} increases steadily and, by 2023–2025, ranks above \emph{healthcare\_pharma}.
\emph{ecommerce\_retail} trends upward and reaches low-to-mid teens by 2025.
\emph{banking\_finance} stays small and drifts down.
\emph{insurance} and \emph{government\_public\_admin} remain in single-digit to low-teens bands, with a late uptick for government.
\emph{social\_media} stays minor throughout.
\emph{Others} expands modestly in the last years.

\subsection{Unique predictors vs. unique DOIs with RDC diversity}
\begin{figure}[htbp]
  \centering
  \includegraphics[width=0.75\linewidth,keepaspectratio]{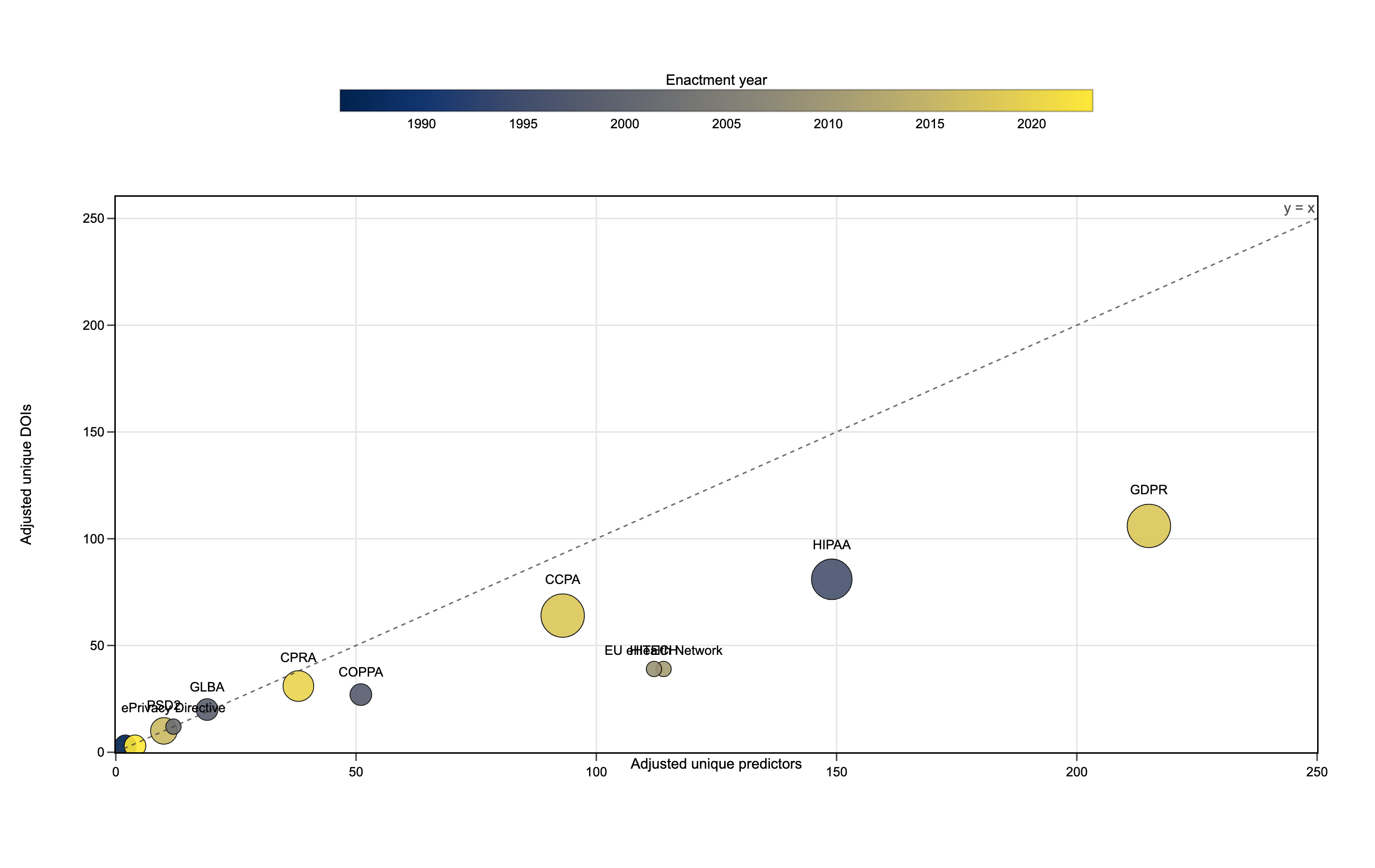}
  \caption{Unique predictors versus unique DOIs per regulation. Point size
           shows RDC diversity. Values on both axes are adjusted.
           Color encodes reference year ($E_r$); y=x reference line shown.}
  \Description{A scatter plot where each point is a regulation, positioned by
               adjusted unique predictors (x) and adjusted unique DOIs (y).
               Larger bubbles indicate more regulated data categories.
               A dotted diagonal line marks y=x. A horizontal colorbar above
               the plot shows the mapping from color to reference year ($E_r$).}
  \label{fig:predictors_vs_dois}
\end{figure}

To understand the relationship between predictor diversity and publication coverage across regulations, we examine how the breadth of unique predictors relates to the number of unique articles discussing them.
Figure~\ref{fig:predictors_vs_dois} presents a scatter plot where each point is a regulation, positioned by unique predictors (x-axis) and unique articles (y-axis).
Marker size encodes RDC diversity; colour encodes reference year ($E_r$).
Labels show regulation names.

The scatter plot reveals distinct positioning of regulations along both dimensions.
GDPR sits top-right with the highest number predictors covered and article counts and the largest bubble,
indicating broad coverage and high class diversity among regulated data categories.
HIPAA is also high on both axes. CCPA is mid-high with substantial diversity.
HITECH and EU eHealth Network cluster mid-right (about 270–290 predictors; about 95–100 articles).
COPPA and CPRA are mid-range (about 90–120 predictors; about 60–80 articles). GLBA is lower-mid (about 45 predictors; about 45 articles).
ePrivacy Directive, PSD2, NIS2, and ECPA are near the origin.


\subsection{Regulated data category distribution by regulation}
\begin{figure}[htbp]
  \centering
  \includegraphics[width=0.75\linewidth,keepaspectratio]{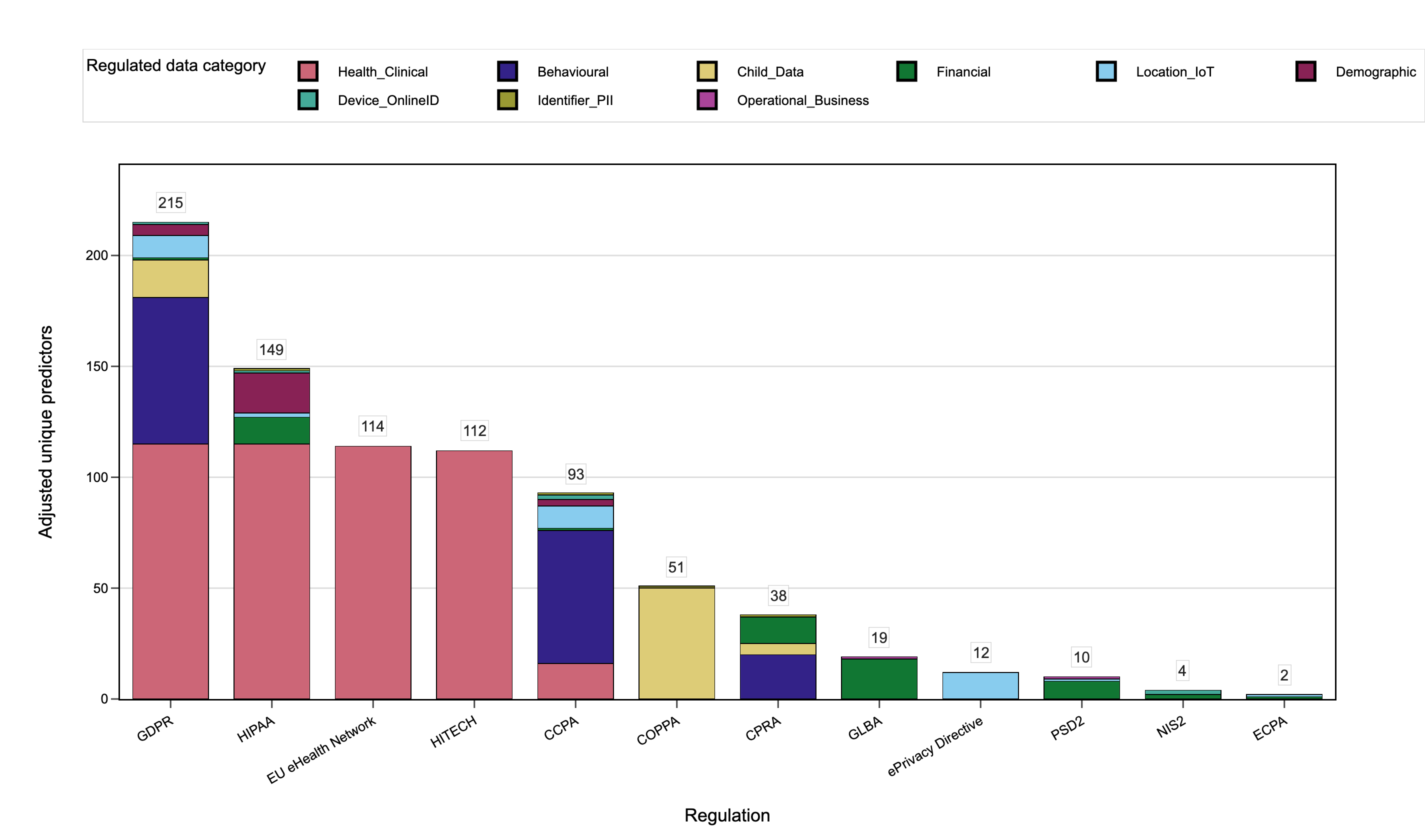}
  \caption{Regulated data category composition by regulation (all). Stacked bars show purity-adjusted counts per category; totals are annotated above each bar.}
  \Description{A stacked vertical bar chart with one bar per regulation. Each bar is partitioned by regulated data category using distinct colors with black outlines. Numeric totals are shown above each bar. X-axis shows regulation identifiers at a \(-45^\circ\) angle. Y-axis shows adjusted counts from zero upward.}
  \label{fig:reg_attr_mix_all}
\end{figure}

Beyond examining the total number of predictors per regulation, understanding the composition of regulated data categories reveals which types of data each regulation primarily governs.
Figure~\ref{fig:reg_attr_mix_all} presents all regulations with bars ordered by adjusted distinct predictors.
We deduplicate by regulation--feature and assign each feature to a single RDC.
A single adjustment is applied per regulation; category stacks are apportioned so that stacked heights equal the per-regulation totals. Totals are annotated above each bar.

The categorical composition reveals regulation-specific patterns.
HIPAA, EU eHealth Network, and HITECH are dominated by Health\_Clinical.
GDPR mixes classes, led by Health\_Clinical and Behavioural, with smaller Child\_Data, Demographic, Location\_IoT, Identifier\_PII, and Device\_OnlineID.
CCPA and CPRA are driven by Behavioural with some Location\_IoT and minor Identifier\_PII and Device\_OnlineID.
COPPA is almost entirely Child\_Data.
GLBA and PSD2 are mainly Financial.
The ePrivacy Directive is mostly Location\_IoT.
NIS2 and ECPA have very small totals.

\subsection{Correlation of regulation–industry profiles}
\begin{figure}[htbp]
  \centering
  \includegraphics[width=0.75\linewidth,keepaspectratio]{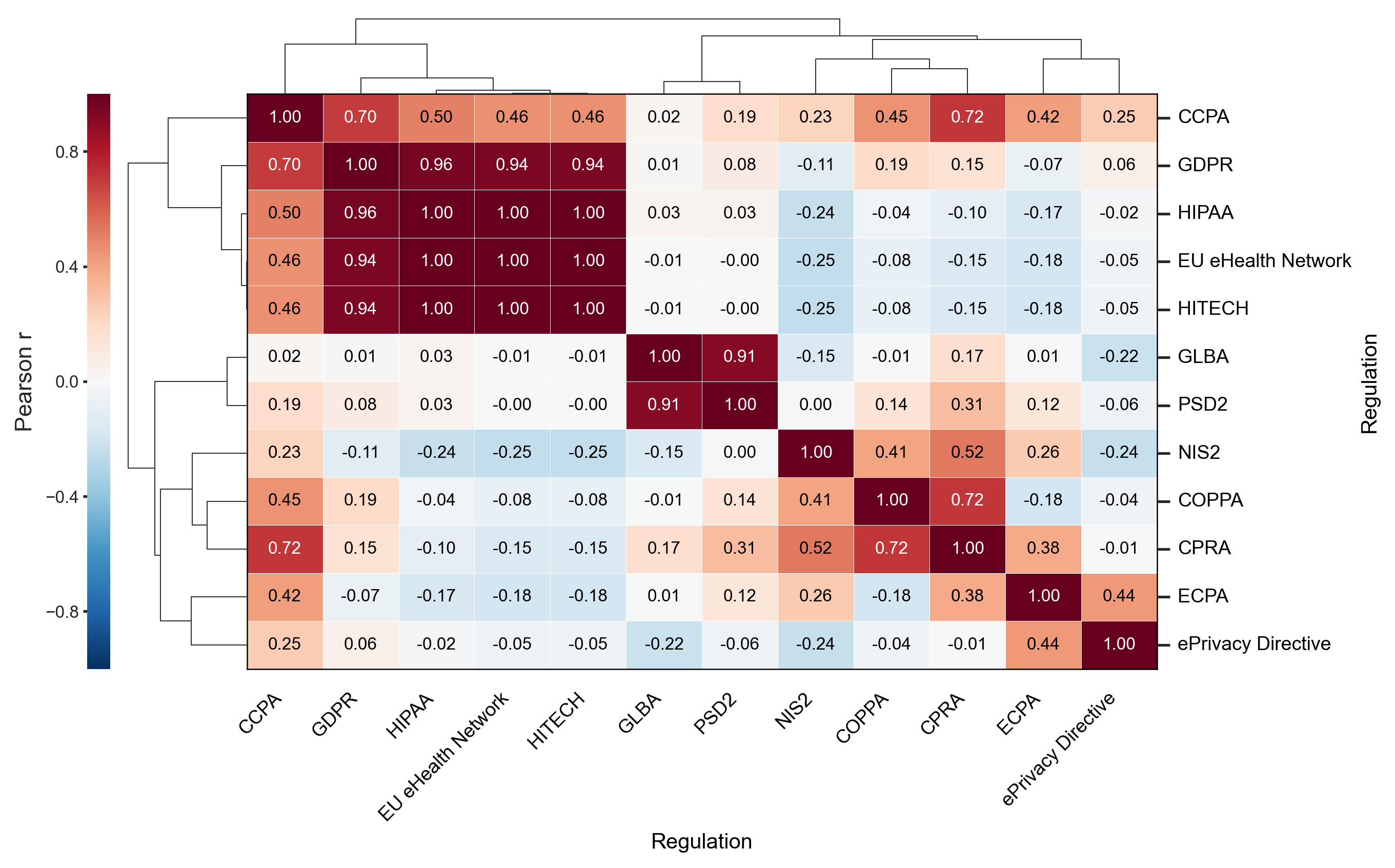}
  \caption{Correlation between regulations based on domain-level distinct-feature
           profiles. Pearson $r$ after purity adjustment; clustered rows and columns.
           Horizontal colorbar shows the $r$ scale.}
  \Description{A clustered correlation heatmap of regulations with a horizontal
               colorbar at the top. Cells are annotated with correlation values.
               A thin black frame surrounds the heatmap. Tick labels are readable
               with regulations on both axes.}
  \label{fig:reg_domain_corr}
\end{figure}

To identify which regulations exhibit similar industry-level patterns of regulated predictor usage, we examine the correlation structure of their domain profiles.
For each regulation, we construct a vector of distinct regulated predictors per industry,
compute Pearson correlations across regulations, and present the results in a clustered heatmap.
Figure~\ref{fig:reg_domain_corr} displays the correlation matrix.

The correlation structure reveals three distinct families (absolute correlation at least 0.60):
(i) finance/telecom/payments cluster—ECPA, GLBA, PSD2;
(ii) broad consumer plus health privacy cluster—GDPR, CCPA, HIPAA, EU eHealth Network, HITECH; and
(iii) education/children/consumer cluster—CPRA, COPPA, FERPA. NIS2 and the ePrivacy Directive show low or negative correlations with most others.
Correlation indicates similarity of industry distributions and does not imply legal overlap.

\subsection{Industry sector × RDC clustered heatmap }
\begin{figure}[htbp]
  \centering
  \includegraphics[width=0.75\linewidth,keepaspectratio]{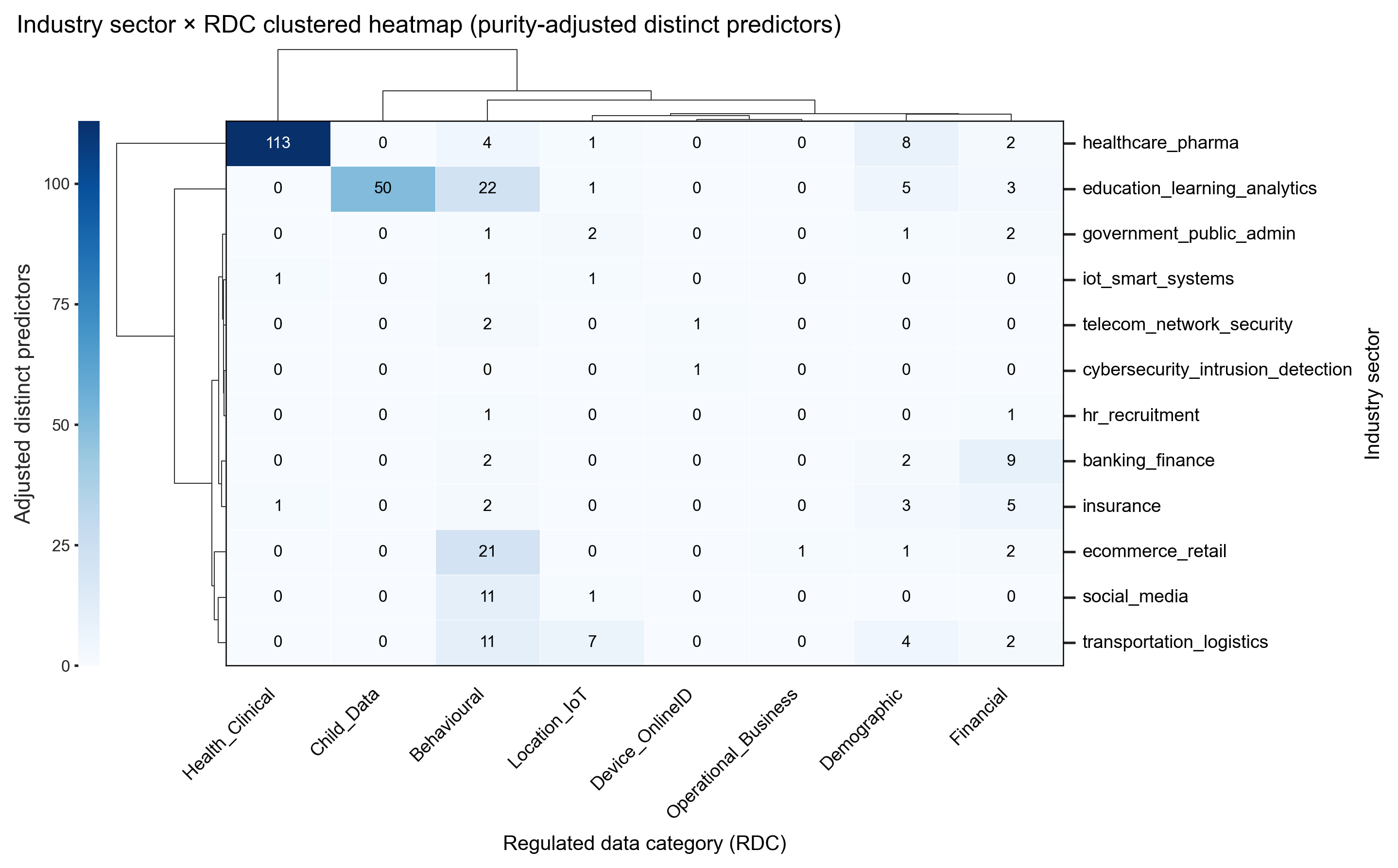}
  \caption{Clustered heatmap of industry sector by regulated data category (RDC).
           Cells show adjusted distinct predictors with annotations.
           Blue palette consistent with the heatmap color scheme used here; the vertical colorbar sits outside
           labels and dendrograms.}
  \Description{A clustered heatmap with industries on rows and regulated data
               categories on columns. A slim vertical blue-scale colorbar sits
               outside the row dendrogram. Each cell is annotated with its
               adjusted count; a thin black frame surrounds the heatmap.}
  \label{fig:domain_by_attr_heatmap}
\end{figure}

To understand how different types of regulated data are distributed across application domains, we examine the joint structure of industry sectors and regulated data categories.
Figure~\ref{fig:domain_by_attr_heatmap} presents counts of distinct regulated predictors for each industry × RDC pair, with hierarchical clustering applied to both rows and columns.

The clustered heatmap reveals several structural patterns.
Examining dominant associations, Healthcare–Pharma aligns strongly with Health–Clinical. Education–Learning Analytics concentrates in Child Data and Behavioural. E-commerce/Retail and Social Media concentrate in Behavioural. Banking/Finance and Insurance concentrate in Financial with secondary Demographic. Transportation–Logistics shows Behavioural and Location/IoT.
The row clustering shows that Healthcare–Pharma and Education form a high-volume block. Banking/Finance clusters with Insurance. Security-oriented domains (Telecom/Network Security, Cybersecurity/Intrusion Detection) group together and are relatively sparse.
Column clustering indicates that Behavioural and Child Data appear near each other. Financial pairs with Demographic. Identifier/PII and Operational Business remain marginal across industries.
Regarding sparsity, Government/Public Administration, HR/Recruitment, IoT/Smart Systems, and security-oriented domains show low counts across most categories.

\subsection{Regulation → RDC coverage}
\begin{figure}[htbp]
  \centering
  \includegraphics[width=0.75\linewidth,keepaspectratio]{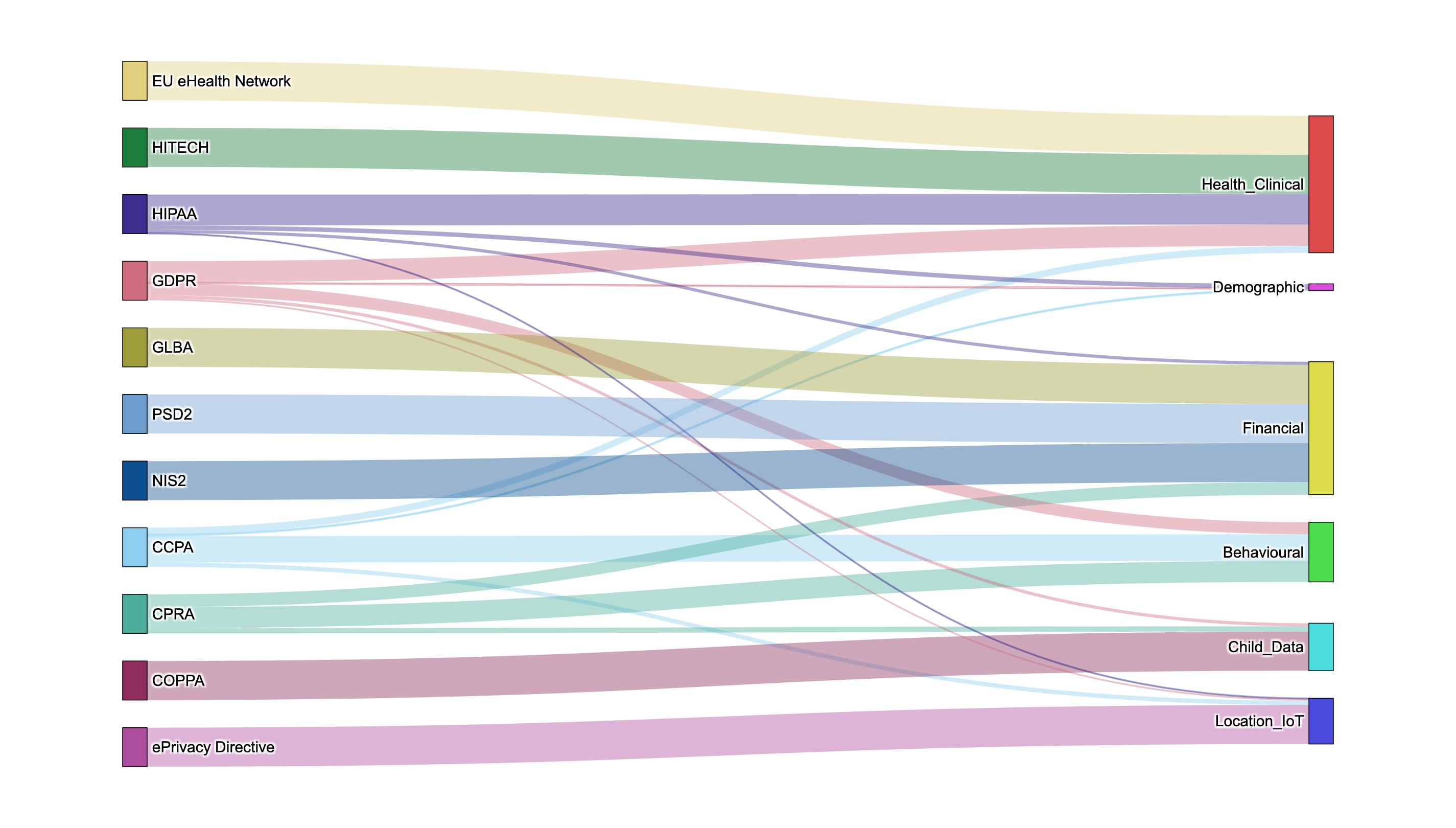}
  \caption{Flows from regulations to regulated data categories (RDCs).
           Link width encodes the proportion of each regulation’s adjusted
           distinct predictors directed to each RDC (normalization per
           regulation). Node colors use Paul~Tol for regulations and Glasbey
           for RDCs; link colors inherit the source regulation hue.}
  \Description{A left-to-right Sankey diagram with regulations as source nodes
               and regulated data categories as target nodes. Links are colored
               by the source regulation with partial transparency. The diagram
               uses white backgrounds and black node outlines for print clarity.}
  \label{fig:reg_to_attr_sankey}
\end{figure}

To visualize how each regulation's coverage is distributed across different data types, we examine the allocation of distinct predictors to regulated data categories.
Figure~\ref{fig:reg_to_attr_sankey} presents a Sankey diagram with regulations on the left and regulated data categories (RDCs) on the right. Link width is proportional to the number of distinct reported predictors that the automated validator labeled \texttt{Regulated} with \texttt{High} confidence for each regulation–RDC pair.

The flow patterns reveal regulation-specific concentrations in particular data categories.
Health-focused frameworks (HIPAA, HITECH, EU eHealth Network) show strong flows into \texttt{Health\_Clinical}.
Financial statutes (GLBA, PSD2 and parts of CPRA) are strongly linked to \texttt{Financial}.
COPPA connects primarily to \texttt{Child\_Data}, while the ePrivacy Directive contributes prominently to \texttt{Location\_IoT}.
Broad consumer privacy laws (GDPR, CCPA, CPRA) display more mixed flows, with noticeable shares in \texttt{Behavioural}, \texttt{Financial}, \texttt{Demographic},
and \texttt{Health\_Clinical}. Overall, the diagram highlights distinct RDC profiles by regulation, with GDPR spanning the widest range of regulated data categories.

\subsection{Regulation × RDC: chi-square test of independence}

To formally assess whether regulations exhibit different patterns of coverage across regulated data categories, we apply a chi-square test of independence.
We build an \(R\times C\) contingency table with \(R\) regulations (rows) and \(C\) regulated data categories (RDCs, columns).
Each cell \(O_{rc}\) counts distinct reported predictors for that regulation–RDC pair (deduplicated by predictor), using purity-adjusted counts.
We test \(H_0\): regulation and RDC are independent, with \(E_{rc}=(\text{row}_r\cdot \text{col}_c)/N\) and \(N=\sum_{r,c}O_{rc}\).
The statistic is \(\chi^2=\sum_{r,c}(O_{rc}-E_{rc})^2/E_{rc}\), \(df=(R-1)(C-1)\).
Effect size is Cramér's \(V=\sqrt{\chi^2/(N\cdot \min(R-1,C-1))}\) \cite{agresti2013categorical,cramer1946mathematical}.
We plot standardized residuals \(z_{rc}=(O_{rc}-E_{rc})/\sqrt{E_{rc}}\) where \(z_{rc}>0\) indicates over-representation and \(z_{rc}<0\) under-representation.
Per-cell two-sided \(p\)-values from \(z_{rc}\) are adjusted by Benjamini–Hochberg FDR \cite{benjamini1995controlling} at \(\alpha=0.05\) over the \(m\) cells with \(E_{rc}\ge 5\); tiles are starred when the adjusted \(p\le 0.05\).

The independence test operates on counts of distinct predictor–regulation pairs (deduplicated by predictor).
A paper can contribute multiple pairs if it reports multiple predictors.
This matches the goal of comparing coverage profiles across regulations rather than paper-level rates.
We examined expected counts returned by \texttt{chi2\_contingency}.
For per-cell inference we excluded cells with \(E_{rc}<5\) from the tested family.
The omnibus test used the asymptotic \(\chi^2\) reference; when adequacy was not met we also computed a Monte--Carlo \(p\)-value as a sensitivity check.

\begin{figure}[htbp]
  \centering
  \includegraphics[width=0.75\linewidth,keepaspectratio]{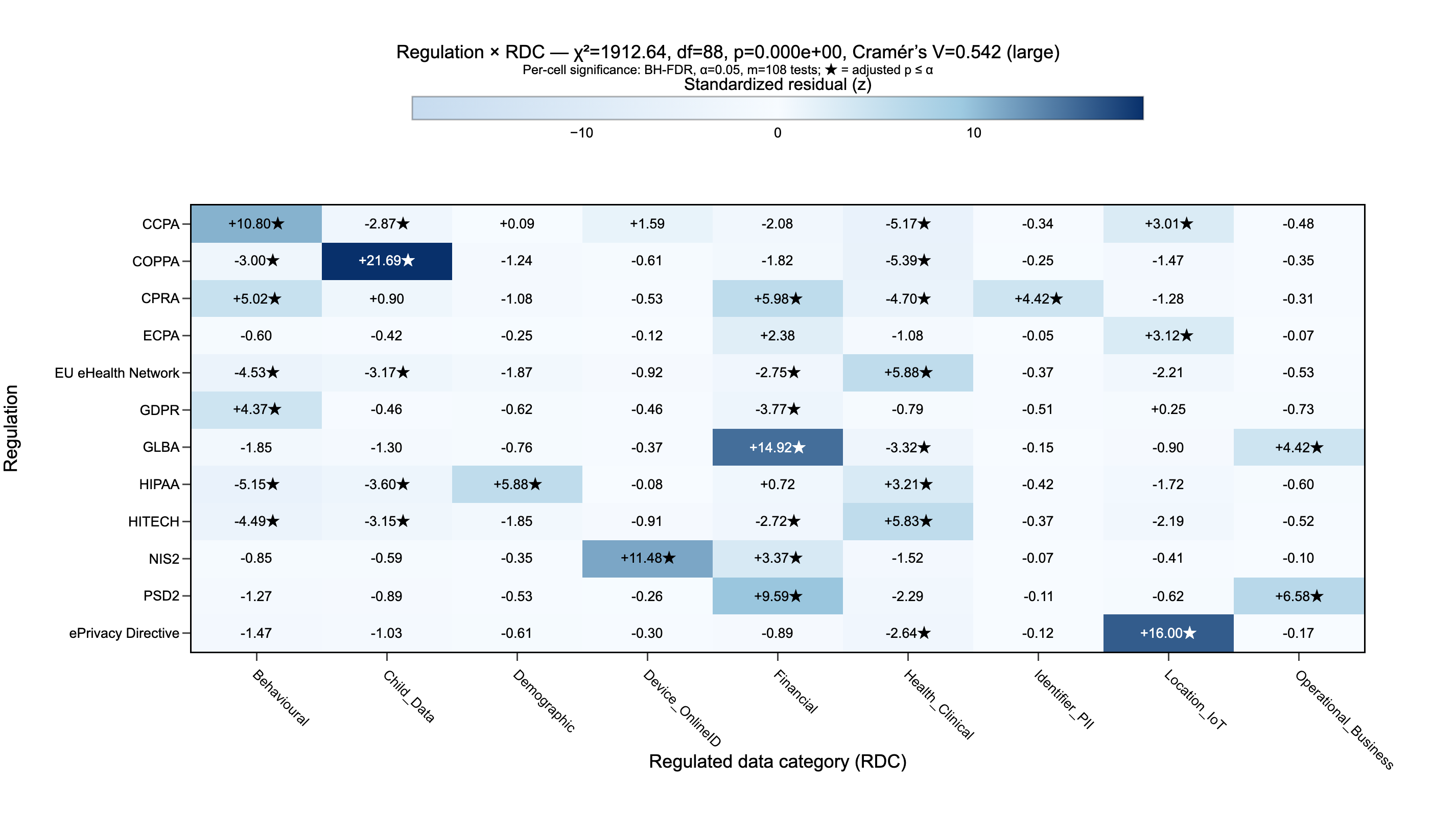}
  \caption{Standardized residuals for regulation–RDC associations using a blue, single-hue diverging palette.
           Near-white denotes \(z\!\approx\!0\); darker blues denote positive residuals; lighter blues denote negative residuals.
           Counts are purity-adjusted.
           Stars mark cells with BH-FDR adjusted \(p\le 0.05\) across the \(m\) tested cells (\(E_{rc}\ge 5\)); \(m\) is shown in the title.
  }
  \Description{Heatmap with regulations on rows and regulated data categories on columns; each tile shows the residual value.
               A horizontal colorbar above the plot. Stars indicate BH-FDR significance at \(\alpha=0.05\).}
  \label{fig:reg_attr_assoc}
\end{figure}

\begin{figure}[htbp]
  \centering
  \includegraphics[width=0.75\linewidth,keepaspectratio]{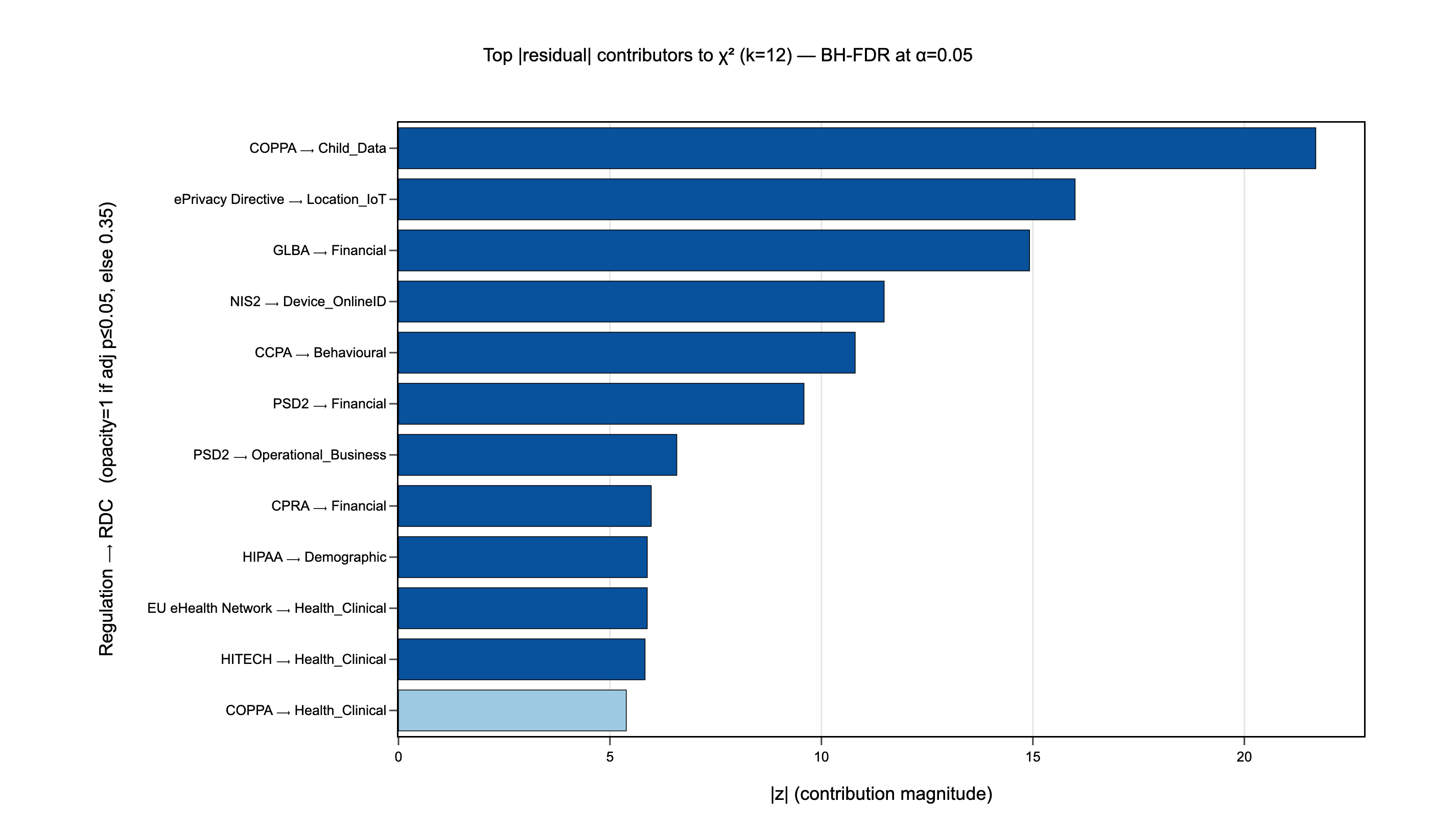}
  \caption{Top \(|\)residual\(|\) contributors to \(\chi^2\) (largest \(|z|\)).
           Bars are colored by sign (dark blue for positive, light blue for negative).
           Opacity encodes BH-FDR significance: opaque if adjusted \(p\le 0.05\), else 0.35.
           Hover data in the interactive figure reports observed adjusted counts and expectations.}
  \Description{Horizontal bar chart of the largest absolute standardized residuals by regulation–RDC pair; opacity shows BH-FDR significance.}
  \label{fig:reg_attr_assoc_top}
\end{figure}

Figures~\ref{fig:reg_attr_assoc} and \ref{fig:reg_attr_assoc_top} present the results.

The omnibus test rejects independence with a large association: \(\chi^2=1912.64\), \(df=88\), \(p\approx 0\), Cramér's \(V=0.542\) (large).
After BH-FDR at \(\alpha=0.05\), starred tiles indicate significant deviations; the companion bar chart shows the largest absolute residuals.
Among the largest positive residuals (over-representation), the following pairs stand out and are starred in Figure~\ref{fig:reg_attr_assoc}:
COPPA\(\rightarrow\)\textbf{Child\_Data};
ePrivacy Directive\(\rightarrow\)\textbf{Location\_IoT};
GLBA\(\rightarrow\)\textbf{Financial};
NIS2\(\rightarrow\)\textbf{Device\_OnlineID};
CCPA\(\rightarrow\)\textbf{Behavioural};
PSD2\(\rightarrow\)\textbf{Financial};
PSD2\(\rightarrow\)\textbf{Operational\_Business};
CPRA\(\rightarrow\)\textbf{Financial};
HIPAA\(\rightarrow\)\textbf{Demographic};
EU eHealth Network\(\rightarrow\)\textbf{Health\_Clinical};
HITECH\(\rightarrow\)\textbf{Health\_Clinical}.
Large negative residuals (under-representation) appear, for example, at COPPA\(\rightarrow\)Health\_Clinical and HIPAA\(\rightarrow\)Behavioural.
These findings indicate that regulations exhibit distinct coverage profiles across RDCs.
Given a feature used in decision trees and its identified regulated data category, practitioners can focus compliance assessment on the subset of regulations showing strong positive associations with that category.
For instance, a feature classified as Health\_Clinical is predominantly governed by HIPAA, HITECH, and EU eHealth Network, while a feature classified as Child\_Data is almost exclusively regulated by COPPA.
Compliance checks should therefore be targeted by category–regulation pairs rather than assuming a uniform mix across laws.

\section{Data, Code, and Artifacts}\label{sec:artifacts}
We provide DOI lists, normalized predictors, class mappings, validator prompts, and analysis
notebooks so that other researchers can inspect our pipeline and reproduce our analyses.
All materials are available in a public repository at
\url{https://github.com/Alex-Veskoukis/data_mining_and_validation_using_AI}.
We package these materials to support artifact evaluation.

\section{Replication Package}
The full replication package, including all code, processed data, and analysis notebooks
described in Section~\ref{sec:artifacts}, is available at
\url{https://github.com/Alex-Veskoukis/data_mining_and_validation_using_AI}.

\section{Societal and Policy Implications}
The measurements in this study show where reported decision-tree practice intersects with privacy law.
Cataloged links between reported predictors and regulated data categories indicate in which parts of the literature
these intersections are more common across sectors and over time. These patterns describe published work, not legal
scope or enforcement outcomes, and they summarize reported practice rather than deployment.

The corpus and supporting materials aim to support transparency and accountability.
Paragraph-level citations to official texts make each coverage decision open to inspection.
The released corpus, code, and prompts allow independent checks of counts and figures.
They also allow reuse of the mapping and audit procedures for other corpora.

The aggregated counts by regulation and regulated data category offer an empirical basis for discussions about disclosure
expectations and documentation practices in applied research. Many deployments adopt techniques first reported in the scholarly literature,
and applied studies often rely on sector datasets and widely used benchmarks. For this reason, the paper-level shares we report
provide a conservative proxy for industry uptake. At the same time, the results are not compliance determinations and do not
replace jurisdiction-specific legal analysis, which must consider context and the full set of applicable rules.


\section{Limitations}
Our results are subject to scope limitations from data selection and study design. Analyses use predictors that authors
explicitly list, so unreported variables are out of scope. The extractor operated on abstracts only under a precision-first strategy,
so studies that use regulated predictors without listing them in the abstract are missed.
Results include only predictor–regulation pairs labeled \texttt{Regulated} with \texttt{High} confidence by the validation step;
pairs that do not reach this status earlier in the pipeline are not reintroduced later. Findings cover decision-tree studies
through 2025 and may not generalize to other model families or to unpublished practice.
Results include only predictor–regulation pairs labeled \texttt{Regulated} with \texttt{High} confidence by the validation step,
so all reported rates are conservative lower bounds under a precision-first design (see Section~\ref{sec:dataset-construction}, \emph{Dataset Construction}).

There are also limitations in how we map predictors to regulation. Coverage decisions are grounded in quoted paragraphs
from official instruments, and we use these paragraph-anchored fragments to decide whether a predictor or regulated d
ata category is covered. Cross-references, implicit coverage, or jurisdiction-specific nuances may therefore be missed.
Evidence concerns published practice rather than deployment, and the study does not offer legal advice;
legal obligations depend on the specific context and jurisdiction.

Finally, the extraction and labeling pipeline introduces residual error.
RDC tags on predictors and on legal excerpts can be wrong or incomplete,
even though we quantify status-only agreement with a stratified audit and apply a one-sided adjustment at reporting.
OCR and parsing of legal PDFs, together with inconsistent naming of predictors and industries, also introduce error.
Large language model (LLM) extraction and classification can under- or over-extract certain predictor types,
for example specific English phrasings or acronyms.
This can bias regulated-category rates and between-group comparisons; schema constraints, versioned prompts, and stratified
human audits mitigate but do not eliminate this risk.

\section{Conclusion}
This paper closes a corpus-scale measurement gap by reporting where published decision-tree studies report predictors that
fall under law-defined data types across sectors and years. The produced counts serve as an informative proxy for how often such
regulated predictors appear in decision-tree applications in industry and applied research.

The results highlight concentrated exposure at the sector level. Healthcare/Pharma accounts for over half of regulated-predictor
records. Education/Learning Analytics grows over time and holds the largest share by 2023–2025, while Transportation/Logistics
peaks in the mid-2010s.

At the regulation level, GDPR and HIPAA are most impactful. GDPR has the broadest cross-sector footprint and class diversity,
and HIPAA dominates health. HITECH and the EU eHealth Network add further depth in healthcare. COPPA concentrates in education.
GLBA is mainly banking/finance. ePrivacy shows a pocket in transportation, while NIS2 and ECPA remain sparse.

At the regulated data category level, \textit{Health\_Clinical} is the top regulated category. \textit{Behavioural} and
\textit{Demographic} follow. \textit{Location\_IoT} and \textit{Financial} appear at moderate levels, while identifier/contact
classes are rare. In timing, regulated-predictor counts cluster in 2020–2022 across multiple frameworks rather than exactly at
regulation's reference year, with GDPR peaking near 2021. Together, these measurements quantify where reported practice is most likely to be legally
governed and which combinations of sector, regulation, and data category carry the greatest exposure.

Given the corpus-level concentration of regulated predictors across sectors and years, privacy-preserving Machine Learning should be treated
as standard practice. It can reduce legal and operational exposure for industry and encourage privacy-enhancing technology (PET)
centred designs and explicit privacy reporting, in line with international AI risk-management principles.

\begin{acks}
Generative AI assisted with extraction and semantic classification; all outputs were schema-constrained and audit-checked.
We release the resulting AI outputs to support reproduction.
\end{acks}

\section*{Conflict of Interest}
The authors declare no competing interests.

\section*{Ethics Statement}
This work used only public article text, metadata, and statute excerpts; no human subjects or personal data were involved.

\section*{Data Availability}
The replication package for this study, including code, processed data, and analysis notebooks, is available at
\url{https://github.com/Alex-Veskoukis/data_mining_and_validation_using_AI}.



\clearpage
\bibliographystyle{ACM-Reference-Format}
\bibliography{refs}

\clearpage
\appendix

\section{Prompt and Validator Contracts}
This appendix records the fixed prompt contracts used in the pipeline.
It starts with an overview and then specifies each step's scope, inputs, outputs, and guardrails.
\subsection*{Overview}
The following contracts specify, for each pipeline step, the scope, inputs, outputs, guardrails, and the exact prompt text.

\subsection{Step 1: Relevance classification}
\paragraph{Scope.} Determine whether a publication presents or applies a decision-tree algorithm using only discipline-agnostic metadata.
The objective is a strict binary filter that keeps the downstream pipeline compact.
\paragraph{Inputs.} Title, abstract, and venue text for one paper as retrieved from metadata aggregators. No full text, figures, or references are consulted.
\paragraph{Outputs.} A single token: \texttt{Relevant} or \texttt{Not relevant}. No explanations are returned to minimize ambiguity and simplify audit trails.
\paragraph{Guardrails.} Inputs are constrained to title, abstract, and venue; outputs are restricted to a binary token.
This isolates the gate from full-text leakage and enforces a deterministic decision surface for parsing and auditing.
Accepted tree synonyms are intentionally narrow so ensemble-only mentions are classified as not relevant, reducing topic noise that would inflate
downstream industry counts and contaminate predictor harvesting.
The gate mirrors corpus construction where deduplicated DOIs branch cleanly into “relevant” or “not,” stabilizing later proportions and sensitivity checks.

\begin{promptblock}{Decision tree - Relevance classification prompt}
You are an expert in privacy-preserving machine learning. Your task is to read a paper’s title,
abstract and venue, and decide whether it presents or applies a decision-tree-based machine
learning algorithm. Respond with exactly one of: 'Relevant' or 'Not relevant', and nothing else.
\end{promptblock}

\subsection{Step 2: Domain assignment}
\paragraph{Scope.} Map each relevant article to exactly one target industry sector to enable stratified analyses and subsequent regulation matching.
\paragraph{Inputs.} Title, abstract, keywords, and venue for one relevant paper.
\paragraph{Outputs.} Exactly one industry string from a controlled vocabulary of 13 options. Multi-labeling is not permitted.
\paragraph{Guardrails.} The classifier is bound to a closed, one-of-13 ontology with an explicit \texttt{none\_of\_the\_above} escape,
aligning with the project’s industry taxonomy and preserving cross-source comparability.
Forcing exactly one string ensures mutual exclusivity and simplifies merges and agreement checks during corpus harmonization.
Using only title, abstract, keywords, and venue keeps evidence consistent with upstream retrieval fields and prevents label bleed from full text.
The label can then be reconciled with the searched industry to form validated strata for analysis.
\begin{promptblock}{Domain classification prompt}
You are an expert in classifying scientific articles by industry. Read a articles’ title, abstract,
keywords, and venue, and choose exactly one of these 13 industry sectors:
1. banking_finance
2. healthcare_pharma
3. insurance
4. ecommerce_retail
5. telecom_network_security
6. social_media
7. education_learning_analytics
8. iot_smart_systems
9. government_public_admin
10. cybersecurity_intrusion_detection
11. hr_recruitment
12. transportation_logistics
13. none_of_the_above
Respond with exactly one industry string from the list above, and nothing else.
\end{promptblock}

\subsection{Step 3: Predictor extraction}
\paragraph{Scope.} Identify only the explicitly named predictors that the abstract states were used to train a decision tree, prioritizing high precision over recall.
\paragraph{Inputs.} Title, venue, abstract, and the industry label from Step~2.
\paragraph{Outputs.} A JSON list where each item contains a short predictor name and the single full sentence from the abstract that explicitly mentions it in the modeling context.
\paragraph{Guardrails.} The prompt prohibits inference and requires one verbatim evidence sentence per predictor,
yielding auditor-ready provenance without re-runs and enabling strict, schema-first ingestion.
Allowing an empty list prevents forced positives when abstracts omit predictors.
Case-insensitive exact matching plus a same-sentence rule blocks cross-sentence grafting,
so only predictors tied to an explicit decision-tree usage remain. 
These constraints align with the data model in which predictors are deduplicated, mapped, and linked to regulations without guessing.
\begin{promptblock}{predictor mining prompt}
You are building a predictor table for decision-tree models.
For the scientific article below, do
the following:

1. Identify each **explicit predictor (predictor or attribute)** used in the decision-tree
   described in the following abstract.
   - A predictor is a variable or attribute that is explicitly mentioned in the abstract as
     being used in the decision-tree model.
   - Do not infer or assume predictors that are not explicitly stated in the abstract.

2. For each predictor, locate the **one full sentence** in the abstract that contains the predictor
   name exactly (case-insensitive substring match).
   - The sentence must explicitly mention the predictor in the context of the decision-tree model.
   - Do not include multiple sentences or paragraphs as evidence. Only return the single sentence
     that mentions the predictor.

3. IMPORTANT: If no predictors (predictors or attributes) are explicitly mentioned in the abstract,
   return an empty list.

Return only this JSON object (no extra text):
{
  "predictors": [
    {
      "name": "<short predictor label—for example, “Age”>",
      "evidence": "<the full quoted sentence from the abstract>"
    }
  ]
}

Scientific article:
<<<
Title: {title}
Venue: {venue}
Abstract: {abstract}
Domain: {industry}
>>>
\end{promptblock}

\subsection{Step 4: Predictor validation}
\paragraph{Scope.} Confirm that every extracted item is truly a predictor used with a decision-tree method, 
not an outcome, generic group, or a model unrelated to decision trees.
\paragraph{Inputs.} Title, abstract, and the predictor list produced by Step~3.
\paragraph{Outputs.} Exactly one string: \texttt{Valid} if all predictors satisfy the rules, otherwise \texttt{Not valid}. No partial credit.
\paragraph{Guardrails.} The validator enforces an all-or-nothing decision over the extracted set and defaults to \texttt{Not valid} when uncertain.
Accepting only single-tree synonyms and rejecting ensemble-only claims prevents upstream overreach from contaminating the training-predictor list.
Disallowing title-only or generic group terms removes frequent false positives that inflate regulated counts later.
The minimal two-class output keeps parsing deterministic and supports reproducibility audits; any format deviation is 
unsalvageable by design, intentionally over-penalizing noisy abstracts to maintain the precision needed for the regulation-link stage.
\begin{promptblock}{predictor validation prompt}
You are a strict validator.

Task
- Validate whether ALL listed predictors are explicitly mentioned in the ABSTRACT as
  predictors used in a decision-tree model.

Output
- Return only one of:
  - "Valid"   -> every listed predictor is explicitly mentioned in the abstract as a predictor
                 in a decision-tree model.
  - "Not valid" -> otherwise.
- Return exactly one of the above strings. No extra text.

Key rules
1) “Explicitly mentioned” means the predictor names appear in the abstract text itself.
   Case-insensitive match is allowed. Minor inflection/plural is allowed. Vague groups like
   “demographics” do NOT count unless each listed predictor is named.
2) “Predictors in a decision-tree model” requires the abstract to state a decision-tree
   method was trained/used with those predictors.
   - Accept synonyms for decision-tree: “decision tree”, “CART”, “ID3”, “C4.5”, “C5.0”,
     “J48”, “CHAID”.
3) If any listed predictor is only implied, only in Title/Keywords, or appears as an OUTCOME/target
   rather than a predictor, return "Not valid".
4) If any information is missing or unclear, default to "Not valid".

Scientific article
<<<
Title: {title}
Abstract: {abstract}
predictors: {predictors}
>>>

Examples
[Example 1 — all predictors present and used with a decision tree -> Valid]
Input:
Title: Predicting Readmission Risk
Abstract: We trained a decision tree to predict readmission using age, prior admissions, and
length of stay...
predictors: ["age", "prior admissions", "length of stay"]
Output:
"Valid"

[Example 2 — one predictor missing -> Not valid]
Input:
Title: Customer Churn Analysis
Abstract: We trained a decision tree using tenure and monthly charges to classify churn...
predictors: ["tenure", "monthly charges", "contract type"]
Output:
"Not valid"

[Example 3 — predictors named but as outcome, not predictors -> Not valid]
Input:
Title: Estimating Age from Voice
Abstract: A decision tree predicts age from acoustic markers...
predictors: ["age"]
Output:
"Not valid"

[Example 4 — only tree-based ensemble mentioned, no explicit decision tree -> Not valid]
Input:
Title: Credit Default Prediction
Abstract: We used a random forest with income and age...
predictors: ["income", "age"]
Output:
"Not valid"

[Example 5 — decision-tree synonym used (C4.5) with explicit predictors -> Valid]
Input:
Title: Hypertension Screening
Abstract: A C4.5 decision tree was trained using BMI and systolic blood pressure...
predictors: ["BMI", "systolic blood pressure"]
Output:
"Valid"

[Example 6 — vague group label in abstract -> Not valid]
Input:
Title: Loan Approval Models
Abstract: We used a decision tree with demographic factors to predict approval...
predictors: ["age", "income"]
Output:
"Not valid"
\end{promptblock}

\subsection{Step 5: Predictor \texorpdfstring{$\rightarrow$}{→} regulated data class mapping}
\paragraph{Scope.} Map one already-validated predictor name to exactly one RDC to enable later regulation matching and roll-ups.
\paragraph{Inputs.} One predictor string.
\paragraph{Outputs.} A minimal JSON object with \texttt{class} and \texttt{rationale} (15 words or fewer). 
No other fields.
\paragraph{Guardrails.} The mapping uses a closed ontology with ordered decision rules and requires a compact rationale, creating a stable bridge from heterogeneous predictor names to regulation-aware categories without post-hoc heuristics. The “exactly one class” rule ensures each predictor contributes to at most one regulatory pathway, simplifying joins with laws and preventing double counting. The brief rationale yields human-checkable justifications while remaining machine-parsable. Ambiguity resolves to \texttt{Other}, quarantining uncertain cases from high-confidence analyses while preserving them for sensitivity checks.
\begin{promptblock}{predictor classification prompt}
Role: Compliance analyst.
Task: Map ONE incoming predictor name to EXACTLY ONE class. Provide a brief rationale (≤15 words).
Output: Return ONLY a JSON object with keys `class` and `rationale`. No extra text.
Do not transform, normalize, or expand the predictor name. If ambiguous, choose Other.

Classes:
1. Identifier_PII (SSN, national_id, passport)
2. Contact_Info (email, phone, messenger handle)
3. Device_OnlineID (device_id, IP, cookie_id, session_id, ad_id, MAC, IMEI)
4. Biometric (fingerprint, face embedding, iris, voiceprint)
5. Location_IoT (GPS, latitude, longitude, home/work address, cell tower)
6. Health_Clinical (diagnoses, labs, ICD codes, medications, vital signs)
7. Financial (income, salary, credit score, card/account numbers, balance)
8. Child_Data (data about minors or pupils)
9. Demographic (age, gender, ethnicity, nationality, marital status)
10. Behavioural (clicks, browsing, purchase history, time-on-page, login frequency)
11. Environmental (weather, temperature, air quality, noise level, light)
12. Operational_Business (SKU, product_id, transaction_id, order_id, process status)
13. Other

Decision rules (apply in order):
1) Biometric term → Biometric.
2) Health/clinical term or code → Health_Clinical.
3) Device/online identifier (IP, cookie, device_id, session_id, ad_id) → Device_OnlineID.
4) Location/address/coordinates/GPS/cell tower → Location_IoT.
5) Government/person identifier (SSN, passport, national_id, tax_id) → Identifier_PII.
6) Contact channel (email, phone, messenger handle) → Contact_Info.
7) Financial value or account/card numbers → Financial.
8) Child/minor/student-specific data → Child_Data.
9) Human attribute like age/gender/race → Demographic.
10) User actions/usage patterns → Behavioural.
11) Ambient conditions/sensors not tied to identity → Environmental.
12) Business/operational artifacts (SKU, order_id, process fields) → Operational_Business.
13) Else → Other.

Ambiguity:
- If the term is an OUTCOME/label, still classify by its data type.
- If the term is generic:
  • 'user_id'/'patient_id'/'customer_id' → Identifier_PII.
  • 'device_id'/'session_id'/'ad_id' → Device_OnlineID.
  • 'order_id'/'transaction_id'/'sku' → Operational_Business.
- If insufficient context, choose Other and state why.

Constraints:
- Exactly one class.
- Rationale ≤15 words. Plain and factual.
- Output must be valid JSON with only `class` and `rationale`.

Examples (inputs → outputs):
predictor='ip_address' → {"class":"Device_OnlineID","rationale":"Network identifier used for online tracking"}
predictor='home_address' → {"class":"Location_IoT","rationale":"Physical location of an individual"}
predictor='age' → {"class":"Demographic","rationale":"Personal attribute describing age"}
predictor='user_id' → {"class":"Identifier_PII","rationale":"Direct identifier of a person"}
predictor='order_id' → {"class":"Operational_Business","rationale":"Business transaction identifier"}
predictor='heart_rate' → {"class":"Health_Clinical","rationale":"Clinical vital sign measurement"}
predictor='credit_card_number' → {"class":"Financial","rationale":"Financial account identifier"}
predictor='student_grade' → {"class":"Child_Data","rationale":"Data concerning a minor’s schooling"}
predictor='click_through_rate' → {"class":"Behavioural","rationale":"User interaction behavior metric"}
predictor='pm2_5' → {"class":"Environmental","rationale":"Ambient air quality measure"}
predictor='email' → {"class":"Contact_Info","rationale":"Direct electronic contact channel"}
predictor='face_embedding' → {"class":"Biometric","rationale":"Unique biometric template"}
predictor='unknown_predictor_x' → {"class":"Other","rationale":"Insufficient information to classify reliably"}
\end{promptblock}

\subsection{Step 6: Legal passage tagging}
\paragraph{Scope.} Scan authoritative legal texts and mark each quoted fragment with one or more RDCs to create a reusable catalog of regulation signals.
\paragraph{Inputs.} Law fragments from curated PDFs and the fixed RDC vocabulary.
\paragraph{Outputs.} A minimal JSON object per fragment with a \texttt{regulated} boolean, a \texttt{classes} list restricted to the allowed names, and a \texttt{rationale} of at most 15 words.
\paragraph{Guardrails.} The schema fixes a boolean and a class list aligned to the same ontology used for predictors, eliminating drift between technical and legal taxonomies.
Working at fragment level prevents over-generalization from whole statutes while capturing explicit category mentions. Returning \texttt{regulated} with a bounded rationale enables later validators to combine multiple fragments without free-form prose parsing; the result is a reusable catalog aligned to the regulation-link table.
\begin{promptblock}{Legal excerpt mining}
You are a legal-compliance analyst. Decide whether the law fragment mentions that any data
element is regulated. If yes, list the exact matching privacy class from the list below (Other
corresponds to anything else that does not match). Respond only with JSON:
{
  "regulated": true|false,
  "classes":   [at least one class name from the RDCs],
  "rationale": "<≤15 words>"
}
Allowed class names: Identifier_PII, Contact_Info, Device_OnlineID, Biometric, Location_IoT,
Health_Clinical, Financial, Child_Data, Demographic, Behavioural, Environmental,
Operational_Business, Other.
\end{promptblock}

\subsection{Step 7: predictor–regulation validation}
\paragraph{Scope.} Decide regulation status for one predictor using only quoted legal fragments keyed by reference labels.
The decision applies at the predictor–law row level and feeds the high-confidence subset.
The step operates after class tagging so explicit class coverage can satisfy the decision target when the exact predictor is not named.
\paragraph{Inputs.} predictor name; its RDC; optional analyst notes; regulatory context provided as raw text or as a dictionary of assembled from the law-fragment index.
\paragraph{Outputs.} Exactly three lines: \texttt{STATUS: Regulated | Not Regulated}, \texttt{CONFIDENCE: High | Medium | Low}, \texttt{RATIONALE: …}.
The rationale ends with a reference suffix: for Regulated, \texttt{refs: <comma-separated ArticleRef(s)>}; for Not Regulated, \texttt{refs: none}.
Parsers enforce format, normalize tokens, and reject or downgrade deviations.
\paragraph{Guardrails.} The validator consumes only the supplied fragments and ignores external sources.
A predictor is Regulated when a fragment explicitly names the predictor term or unambiguously names its entire RDC with regulatory effect (for example, prohibitions, conditions, consent, or safeguards). Generic governance clauses without predictor/class terms do not qualify. High confidence requires the exact predictor or explicit class term with clear coverage; Medium applies to strong implication; Low applies to ambiguous phrasing or carve-outs. Ties resolve to Not Regulated. Multiple fragments are allowed; when several fragments collectively satisfy coverage, the rationale must enumerate the decisive paragraph-level references drawn from the provided context. Each listed reference must map one-to-one to a quoted paragraph; mismatches or missing paragraph-level identifiers fail high-confidence admission. The rationale is limited to 40 words to preserve traceability and machine-parsability.
\begin{promptblock}{Regulated predictor validation system prompt}
You are a legal analyst.
Decide if ONE predictor is regulated by the PROVIDED regulatory text only.
Do not use outside knowledge.
Ignore titles, keywords, or abstracts not included here.

Decision target
- Is the predictor "{predictor_name}" regulated EITHER by explicit mention of:
  a) the exact predictor name, OR
  b) its whole RDC "{attribute_class}" (or unambiguous legal synonyms of that class),
  within the quoted regulatory text?

Output format (exactly three lines, no extra text)
STATUS: Regulated | Not Regulated
CONFIDENCE: High | Medium | Low
RATIONALE: ≤40 words. State decisive phrase(s). If Regulated, END with the exact refs that
regulate it: <comma-separated ArticleRef(s)>. If Not Regulated, END with refs: none

Rules
1) “Regulated” when the text clearly covers the exact predictor OR clearly covers the whole
   class it belongs to.
   - Coverage includes prohibitions, restrictions, consent requirements, safeguards, or
     processing conditions.
   - If the text lists the class by name (e.g., biometric data) the predictor in that class
     counts as regulated.
2) “Not Regulated” when the text does NOT clearly mention the predictor or its class.
   - Generic principles (lawful basis, transparency, security) without predictor/class do NOT suffice.
   - If the text EXPLICITLY excludes the predictor/class, return “Not Regulated”.
3) Confidence:
   - High: exact predictor OR explicit class term appears (or an unambiguous legal synonym)
     with clear coverage.
   - Medium: close paraphrase or category term strongly implies coverage, but wording is
     less direct.
   - Low: ambiguous wording, weak implication, or conflicting clauses.
4) Ties go to “Not Regulated”. If unsure, choose “Not Regulated” with Low confidence.
5) Use only the quoted regulatory text. Notes are context, not authority.

Examples
Example 1 — Regulated (exact class term present)
STATUS: Regulated
CONFIDENCE: High
RATIONALE: Text covers “biometric data” and restricts processing; fingerprint pattern is biometric.
refs: GDPR Art.9(1)

Example 2 — Not Regulated (no predictor/class)
STATUS: Not Regulated
CONFIDENCE: High
RATIONALE: Record confidentiality only; no hearing metrics or Health_Clinical class. refs: none

Example 3 — Regulated (explicit predictor)
STATUS: Regulated
CONFIDENCE: High
RATIONALE: “IP address” listed as personal data under processing limits. refs: GDPR Recital 30

Example 4 — Not Regulated (ambiguous/vague)
STATUS: Not Regulated
CONFIDENCE: Low
RATIONALE: Only generic ‘personal information’; no clear link to {predictor_name} or {attribute_class}.
refs: none
\end{promptblock}

\paragraph{Prompt — user message template.}
\begin{promptblock}{Regulated predictor validation user prompt}
predictor TO VALIDATE: {predictor_name}
predictor ATTRIBUTE CLASS: {attribute_class}
Notes: {notes}

REGULATORY CONTEXT (use ONLY this text):
{regulatory_context}

QUESTION: Is the predictor "{predictor_name}" regulated either specifically or via its whole class
"{attribute_class}" according to this regulatory context?
\end{promptblock}

\section{Regulation Reference Year Timeline}\label{sec:regulation-enactment-timeline}

\begin{table}[H]   
  \centering
  \small
  \begin{tabular}{@{} l c @{}}
    \hline
    \textbf{Regulation} & \textbf{Reference year ($E_r$)} \\
    \hline
    HIPAA & 1996 \\
    HITECH & 2009 \\
    CCPA & 2018 \\
    CPRA & 2020 \\
    GDPR & 2018 \\
    ePrivacy Directive & 2002 \\
    NIS2 & 2023 \\
    PSD2 & 2016 \\
    EU eHealth Network & 2011 \\
    GLBA & 1999 \\
    COPPA & 1998 \\
    FERPA & 1974 \\
    ECPA & 1986 \\
    \hline
  \end{tabular}
  \caption{Reference years ($E_r$) used in figures and time-series models. Default rule: statute enactment/adoption; EU instruments: adoption or entry-into-force. Where a different milestone is used, it is noted in text.}
  \label{tab:reg_enactment}
\end{table}

\section{AI Model Metadata}\label{app:ai-model-gpt41}
This appendix documents the identification and lifecycle metadata for the language model used in the corpus construction pipeline.

\subsection*{Model identification}
\begin{table}[H]
  \centering
  \small
  \begin{tabular}{@{} l l @{}}
    \hline
    \textbf{Field} & \textbf{Value} \\
    \hline
    Model name & gpt-4.1 \\
    Model version & 2025-04-14 \\
    Lifecycle status & Generally Available \\
    Date created & 11 Apr 2025, 03{:}00 \\
    Date updated & 11 Apr 2025, 03{:}00 \\
    Planned retirement & 11 Apr 2026, 03{:}00 \\
    \hline
  \end{tabular}
  \caption{Core identification and lifecycle details for the language model used.}
  \label{tab:model-metadata-gpt41}
\end{table}

\end{document}